\pdfoutput=1

\documentclass[11pt,usenames,dvipsnames]{article}

\usepackage{EMNLP2023}  

\usepackage{times}
\usepackage{latexsym}
\usepackage[export]{adjustbox}

\usepackage[T1]{fontenc}

\usepackage[utf8]{inputenc}

\usepackage{microtype}
\usepackage{subcaption}
\usepackage{inconsolata}

\usepackage{rotating}
\newcommand\tabrotate[1]{\rotatebox[origin=c]{90}{#1}}

\usepackage{xcolor,colortbl}

\definecolor{Gray}{gray}{0.94}
\definecolor{LightCyan}{rgb}{0.88,1,1}
\newcolumntype{a}{>{\columncolor{Gray}}c}
\newcolumntype{o}{>{\columncolor{white}}c}

\usepackage{soul}

\definecolor{champagne}{cmyk}{0, 0.1608, 0.3529, 0.1}
\definecolor{blond}{cmyk}{0.0078, 0, 0.2863, 0.1}
\definecolor{teagreen}{cmyk}{0.2078, 0, 0.251, 0.1}
\definecolor{celeste}{cmyk}{0.3922, 0.0353, 0, 0.1}
\definecolor{babyblue}{cmyk}{0.3725, 0.2314, 0, 0.1}
\definecolor{lavenderblue}{cmyk}{0.2588, 0.302, 0, 0.1}
\definecolor{brilliantlavender}{cmyk}{0, 0.2235, 0, 0.1}
\definecolor{ilcolor}{cmyk}{0, 0.11, 0.92, 0.06}

\newcommand{\narrowtextsc}[1]{\textls[-50]{\textsc{#1}}}

\newcommand{\lm}[1]{\texttt{#1}}
\newcommand{\sys}[1]{\narrowtextsc{#1}}
\newcommand{\data}[1]{\textsf{#1}}

\newcommand{\ilang}{\sys{InterroLang}\xspace}
\newcommand{\ttm}{\sys{TTM}\xspace}

\usepackage{amsthm}
\usepackage{amsmath}
\usepackage{amssymb}
\usepackage{emoji}
\usepackage{booktabs}
\usepackage{graphicx}
\usepackage{enumitem}
\usepackage{multirow}
\usepackage{xspace}
\usepackage{dsfont}

\newcommand{\present}{$\blacksquare$}

\title{\ilang: \\ Exploring NLP Models and Datasets \\through Dialogue-based Explanations}

\newcommand{\affilsup}[1]{\rlap{\textsuperscript{\normalfont#1}}}
\author{
    Nils Feldhus\affilsup{1}
    \qquad
    \textbf{Qianli Wang}\affilsup{2,1}
    \qquad
    Tatiana Anikina\affilsup{1,3}
    \\
    \textbf{Sahil Chopra}\affilsup{3,1}
    \qquad
    \textbf{Cennet Oguz}\affilsup{1,3}
    \qquad
    \textbf{Sebastian M\"oller}\affilsup{2,1}
    \\
    $^1$German Research Center for Artificial Intelligence (DFKI) \\
    $^2$Technische Universit\"at Berlin, Germany \\
    $^3$Saarland Informatics Campus, Saarbrücken, Germany \\
    \texttt{\{firstname.lastname\}@dfki.de} \\
}

\begin{document}
\maketitle
\begin{abstract}
While recently developed NLP explainability methods let us open the black box in various ways \cite{madsen-2022-post-hoc},
a missing ingredient in this endeavor is an interactive tool offering a conversational interface. 
Such a dialogue system can help users explore datasets and models with explanations in a contextualized manner, e.g. via clarification or follow-up questions, and through a natural language interface.
We adapt the conversational explanation framework \sys{TalkToModel} \cite{slack-2022-talktomodel} to the NLP domain,
add new NLP-specific operations such as free-text rationalization,
and illustrate its generalizability on three NLP tasks (dialogue act classification, question answering, hate speech detection). 
To recognize user queries for explanations, we evaluate fine-tuned and few-shot prompting models and implement a novel Adapter-based approach. 
We then conduct two user studies on 
(1) the perceived correctness and helpfulness of the dialogues, and 
(2) the simulatability, i.e. how objectively helpful dialogical explanations are for humans in figuring out the model's predicted label when it's not shown.
We found rationalization and feature attribution were helpful in explaining the model behavior. 
Moreover, users could more reliably predict the model outcome based on an explanation dialogue rather than one-off explanations.
\end{abstract}

\textit{\footnotesize Disclaimer: This paper contains material that is offensive or hateful.}

\section{Introduction}
\label{sec:intro}

\begin{figure}[ht!]
\centering
\includegraphics[width=\linewidth]{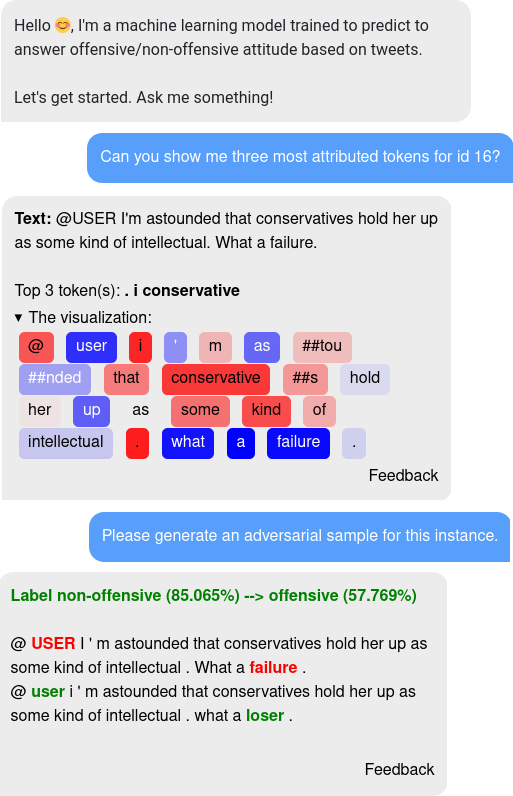}
\caption{\ilang dialogue with token-level attribution and adversarial example operations on a hate speech detection task (\data{OLID}).
Users are aware of IDs in the data, since we provide a dataset viewer (not shown).
}
\label{fig:ilang-interface}
\end{figure}

\begin{figure*}[ht!]
    \centering
    \resizebox{\textwidth}{!}{%
        \includegraphics{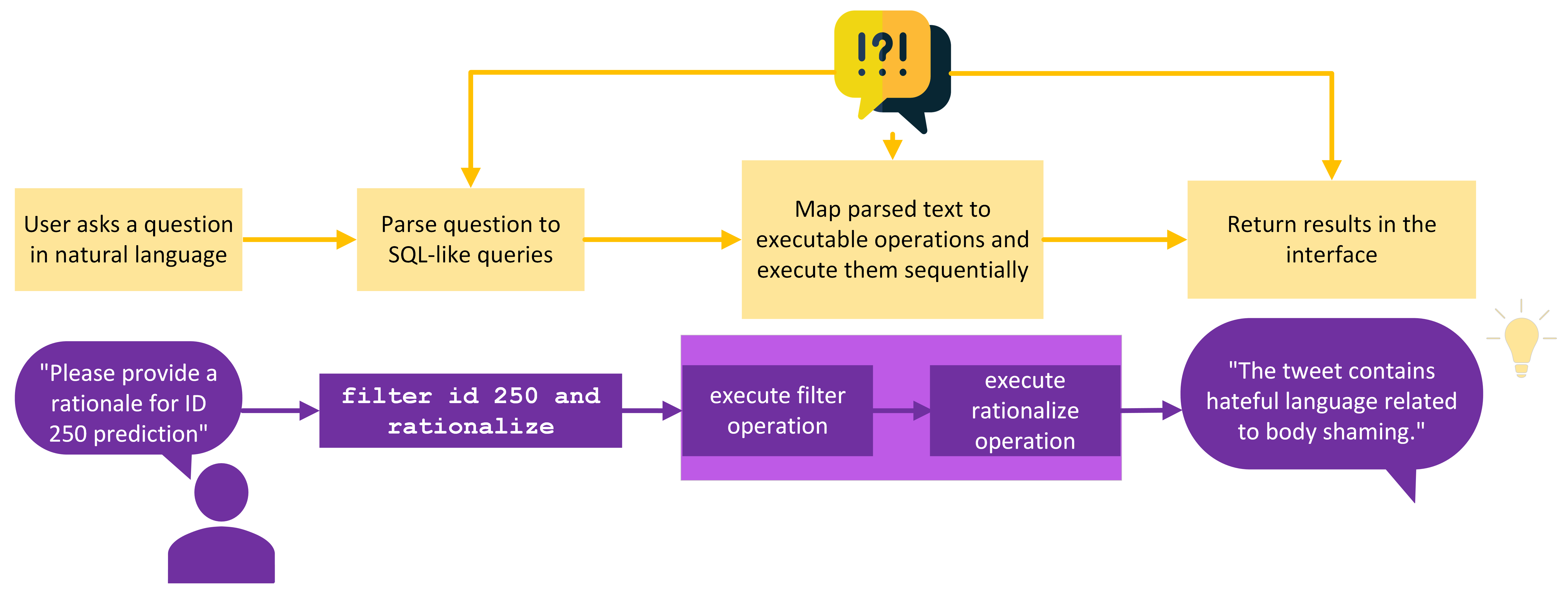}
    }
    \caption{Illustration of how natural language queries from users are parsed into executable operations and their results are inserted in \ilang responses presented through a dialogue interface.
    }
    \label{fig:pipeline}
\end{figure*}

Framing explanation processes as a dialogue between the human and the model has been motivated in many recent works from the areas of HCI and ML explainability \cite{miller-2019-explanation,lakkaraju-2022-rethinking,feldhus-2022-mediators,hartmann-2022-xaines,weld-bansal-2019-crafting,jacovi-2023-diagnosing}.
With the growing popularity of large language models (LLMs), the research community has yet to present a dialogue-based interpretability framework in the NLP domain that is both capable of conveying faithful explanations\footnote{
While it might be tempting to use ChatGPT, we point out the black-box nature of proprietary software: Most interpretability methods require access to gradients, parameters or training data to make faithful explanations of their behavior. Lastly, it is not possible yet to connect other ML models to it for generating explanations.}
in human-understandable terms and is generalizable to different datasets, use cases and models.

One-off explanations can only tell a part of the overall narrative about why a model ``behaves'' a certain way. Saliency maps from feature attribution methods can explain the model reasoning in terms of what input features are important for making a prediction \cite{feldhus-2023-smv}, while counterfactuals and adversarial examples show how an input needs to be modified to cause a change in the original prediction \cite{wu-2021-polyjuice}. Semantic similarity and label distributions can shed a light on the data which was used to train the model \cite{shen-2023-convxai}, while rationales provide a natural language justification for a predicted label \cite{wiegreffe-etal-2022-reframing}.
These methods do not allow follow-up questions to clarify ambiguous cases, e.g. a most important token being a punctuation (Figure~\ref{fig:ilang-interface}) (cf. \citealt{schuff-2022-human}), or build a mental model of the explained models.

In this work, we build a user-centered, dialogue-based explanation and exploration framework, \ilang, for interpretability and analyses of NLP models.
We investigate how the \sys{TalkToModel} (TTM, \citealt{slack-2022-talktomodel}) framework can be implemented in the NLP domain: Concretely, we define NLP-specific operations based on the aforementioned explanation types. Our system, \ilang, allows users to interpret and analyze the behavior of language models interactively.
We demonstrate the generalizability of \ilang on three case studies -- dialogue act classification, question answering, hate speech detection -- for which we evaluate the intent recognition (parsing of natural language queries) capabilities of both fine-tuned (\lm{FLAN-T5}, \lm{BERT} with Adapter) and few-shot LLM (\lm{GPT-Neo}). We find that an efficient Adapter setup outperforms few-shot LLMs, but that this task of detecting a user's intent is far from being solved.
In a subsequent human evaluation (\S \ref{sec:humeval}), we first collect subjective quality assessments on each response about the explanation types regarding four dimensions (correctness, helpfulness, satisfaction, fluency). We find a preference for mistakes summaries, performance metrics and free-text rationales.
Secondly, we ask the participants about their impressions of the overall explanation dialogues. All of them were deemed helpful, although some (e.g., counterfactuals) have some potential for improvement.
Finally, a second user study on simulatability (human forward prediction) provides first evidence for how various NLP explanation types can be meaningfully combined in dialogical settings. Attribution and rationales resulted in very high simulation accuracies and required the least number of turns on average, revealing a need for a longer conversation than single-turn explanations. 
We open-source our tool\footnote{\url{https://github.com/DFKI-NLP/InterroLang}} that can be extended to other models and NLP tasks alongside a dataset collected during the user studies including various operations and manual annotations for the user inputs (parsed texts):
Free-text rationales and template-based responses for the decisions of NLP models include explanations generated from interpretability methods, such as attributions, counterfactuals, and similar examples.


\begin{table*}[ht!]
    \centering

    \resizebox{\textwidth}{!}{%
        \begin{tabular}{ll|l}

        \toprule
        & \textsf{OLID} example instance:  & \textcolor{black!50}{\textit{ibelieveblaseyford is liar she is fat ugly libreal snowflake}} \\
        & & \textcolor{black!50}{\textit{she sold her herself to get some cash !!}} \\
        & & \textcolor{black!50}{\textit{From dems and Iran ! Why she spoke after JohnKerryIranMeeting ?}} \\

        \bottomrule 
        & \textbf{Operation} & \textbf{Desc}ription; \textbf{Q}uestion + \textbf{E}xplanation example \\
        
        \midrule
        \multirow{6}{*}{\centering \rotatebox[origin=c]{90}{\textbf{Attribution}}}
        & \texttt{\textcolor{blue}{nlpattribute}(instance, }
          & \textbf{Desc:} Feature importances on \texttt{instance} at (token~|~sentence)-level \\
        & \texttt{granularity)*}
          & \textbf{Q:} Which tokens are most important? \\
        & & \textbf{E:} \textcolor{black!70}{\textit{fat}}, \textcolor{black!70}{\textit{ugly}} and \textcolor{black!70}{\textit{liar}} are most important for the hate speech label. \\

        & \texttt{\textcolor{blue}{globaltopk}(dataset, k,}
          & \textbf{Desc:} Top \texttt{k} most attributed tokens across the entire \texttt{dataset} \\
        & \texttt{classes)} & \textbf{Q:} What are the three most important keywords for the hate speech label in the data? \\
        & & \textbf{E:} \textcolor{black!70}{\textit{dumb}}, \textcolor{black!70}{\textit{fucking}}, and \textcolor{black!70}{\textit{ugly}} are the most attributed for the hate speech label. \\

        \midrule
        \multirow{9}{*}{\centering \rotatebox[origin=c]{90}{\textbf{Perturbation}}}
        & \texttt{\textcolor{blue}{nlpcfe}(instance,}
          & \textbf{Desc:} Gets \texttt{number} natural language counterfactual explanations for a single \texttt{instance} \\
        & \texttt{number)} & \textbf{Q:} How do you flip the prediction? \\
        & & \textbf{E:} By replacing \textcolor{black!70}{\textit{liar}}, \textcolor{black!70}{\textit{fat}}, \textcolor{black!70}{\textit{ugly}} with neutral nouns and adjectives. \\

        & \texttt{\textcolor{blue}{adversarial}(instance)}
        & \textbf{Desc:} Gets \texttt{number} adversarial examples for a single \texttt{instance} \\
        & & \textbf{Q:} What is the minimal change needed to cause a wrong prediction? \\
        & & \textbf{E:} \textcolor{black!70}{\textit{I question the timing of Dr. Ford's statement following the \#JohnKerryIranMeeting [...]}} \\

        & \texttt{\textcolor{blue}{augment}(instance)}
          & \textbf{Desc:} Generate similar \texttt{instance} \\
        & & \textbf{Q:} Can you generate one more example like this? \\
        & & \textbf{E:} \textcolor{black!70}{\textit{I'm skeptical of her integrity and perceive her as a figure manipulated by political agendas.}} \\

        \midrule
        \multirow{3}{*}{\centering \rotatebox[origin=c]{90}{\textbf{Rat.}}}
        & \texttt{\textcolor{blue}{rationalize}(instance)}
          & \textbf{Desc:} Explain an \texttt{instance} (prediction) in natural language (rationale generation) \\
        & & \textbf{Q:} In natural language, why is this text hateful? \\
        & & \textbf{E:} The text includes multiple instances of insults related to body shaming. \\

        \midrule
        \multirow{6}{*}{\centering \rotatebox[origin=c]{90}{\textbf{NLU}}}

        & \texttt{\textcolor{blue}{keywords}(dataset,}
          & \textbf{Desc:} Show most frequent keywords in the dataset \\
        & \texttt{number)} & \textbf{Q:} What are the most frequent keywords in the dataset? \\
        & & \textbf{E:} \textcolor{black!70}{\textit{USA}}, \textcolor{black!70}{\textit{president}}, \textcolor{black!70}{\textit{democrats}} \\

        & \texttt{\textcolor{blue}{similar}(instance,}
          & \textbf{Desc:} Gets \texttt{number} of training data instances most similar to the current one \\
        & \texttt{number)*} & \textbf{Q:} What is an instance in the data very similar to this one? \\
        & & \textbf{E:} \textcolor{black!70}{\textit{@USER How is she hiding her ugly personality. She is the worst.}} \\
        
        \bottomrule
        \end{tabular}
    }
    \caption{Set of \ilang operations. Descriptions and exemplary question-explanation pairs are added for the hate speech detection use case (\data{OLID}). Operations marked with (*) provide support for custom input instances received from users. This applies to single instance prediction as well (Table~\ref{tab:ttmops}).
    }
    \label{tab:operations}
\end{table*}

\section{Methodology}
\label{sec:method}

\sys{TalkToModel} \cite{slack-2022-talktomodel} is designed as a system for open-ended natural language dialogues for comprehending the behavior of ML models for tabular datasets (including only numeric and categorical features). 
Our system \ilang retains most of its functionalities:
Users can ask questions about many different aspects and slices of the data alongside predictions and explanations. 
\ilang has three main components (depicted in Figure~\ref{fig:pipeline}):
A \textit{dialogue engine} parses user inputs into an SQL-like programming language using either Adapters for intent classification or LLM that treats this task as a seq2seq problem, where user inputs are the source and the parses are the targets.
An \textit{execution engine} runs the operations in each parse and generates the natural language response.
A \textit{text interface} (Figure~\ref{fig:interface}) lets users engage in open-ended dialogues and offers pre-defined questions that can be edited. This reduces the users' workload to deciding on what to ask, essentially.

\subsection{Operations}
\label{sec:operations}

We extend the set of operations in \ttm (App.~\ref{app:ttmops}), e.g. feature attribution and counterfactuals, towards linguistic questions, s.t. they can be used in NLP settings and on Transformers. 
In Table~\ref{tab:operations}, we categorize all \ilang operations into Attribution, Perturbation, Rationalization, and Data.

\paragraph{Attribution}

Feature attribution methods can quantify the importance of input tokens \cite{madsen-2022-post-hoc} by taking the final predictions and intermediate representations of the explained model into account.
Next to simple token-level attributions, 
we can aggregate them on sentence-level or present global top~$k$ attributed tokens across the entire dataset \cite{ronnqvist-etal-2022-explaining}.

\paragraph{Perturbation}
Perturbation methods come in many forms and have different purposes: We propose to include counterfactual generation, adversarial attacks and data augmentation as the main representatives for this category. While counterfactuals aim to edit an input text to cause a change in the model's prediction \cite{wu-2021-polyjuice}, adversarial attacks are about fooling the model to not guess the correct label \cite{ebrahimi-2018-hotflip}. Data augmentation replaces spans in the input, keeping the outcome the same \cite{ross-etal-2022-tailor}.

\paragraph{Rationalization}
Generating free-text rationales for justifying a model prediction in natural language has been a popular task in NLP \cite{camburu-2018-esnli,wiegreffe-etal-2022-reframing}. Such natural language explanations are usually generated by either concatenating the input text with the prediction and then prompting a model to explain the prediction, or by jointly predicting and rationalizing. However, the task has not yet been explored within dialogue-based model interpretability tools.

\paragraph{Similarity}
Inspired by influence functions \cite{koh-liang-2017-influence}, this functionality returns a number of instances from the training data that are related to the (local) instance in question. Since influence functions are notoriously expensive to compute, as a proxy, we instead compute the semantic similarity to all other instances in the training data and retrieve the highest ranked instances.

\subsection{Intent recognition}
\label{sec:intent}

We follow \ttm and write pairs of utterances and SQL-like parses that can be mapped to operations (Table~\ref{tab:operations}) as well as templates that can be filled.

We propose a novel Adapter-based solution \cite{houlsby-2019-parameter,pfeiffer-2020-adapterhub} for intent recognition and train a model which can classify intents representing the \ilang operations (e.g., \texttt{adversarial}, \texttt{counterfactual}, etc.).
We also train a separate Adapter model for the slot tagging, s.t. for each intent we can label the relevant slots. The slot types that can be recognized by the model include \texttt{id, number, class\_names, data\_type, metric, include\_token and sentence\_level}. The training details of the Adapter-based approach are listed in Table \ref{app:adapters-training-params}.
\footnote{Some of the slots are crucial for the intent interpretation and cannot be omitted (e.g., \texttt{id} for the \texttt{show} operation) while other slots are optional and if not specified by the user the default value is chosen. We also implement additional checks for the case when the user input includes deictic expressions (e.g., ``this'' in ``show me a counterfactual for this sample'') in which case the ID of the previous instance is selected.}

The training data for intents are generated from the same prompts that are used for baselines (\lm{GPT-Neo} and \lm{FLAN-T5-base}) with the slot values randomly replaced by the actual values from the datasets (e.g., IDs, class names etc.). Some of the prompts are paraphrased to obtain more diverse training data. Adapter models for intents and slots are fine-tuned on top of the same \lm{bert-base-uncased} model.
The performance of this approach is compared to the prompt-based solution in Table \ref{tab:accuracy}.

\subsection{Dialogue management}
We add dialogue management in the form of parsing consecutive operations (Figure~\ref{fig:pipeline}) and extend it with the ability to handle custom inputs and clarification questions.

\ttm, after translating user utterances into a grammar of production rules, composes its results in a template-filling manner while ensuring semantic coherence between multiple operations. They further argue that such a response generation approach prevents hallucinations commonly found in neural networks and conversational models \cite{dziri-2022-origin-hallucinations}.
However, it makes the dialogue less natural. That is why we also add a range of pre-defined responses for fallback that are chosen at random when applicable. Moreover, the \lm{GPT}-based rationales are also the first example of a fully model-generated response. Our system also recognizes when the user just wants to acknowledge the bot's response or intends to finish the conversation and it generates the appropriate responses (see App. \ref{app:dialogue-example} for an example).

When designing dialogue systems, the task of keeping track of the dialogue history is essential to better inform the selection of the next action or response. Thus, we store the previous operations and ids and can resolve deictic expressions like ``this sample'' or ``it'' to the ID of the previously  mentioned instance. We also check the prediction scores of the intent recognition module to see if there is some problem interpreting the user input, e.g., if several intents get very high scores \ilang asks a clarification question to disambiguate between operations. Also, if we have an intent but some of its non-default slots are missing (not recognized) we can generate a clarification question to resolve it, e.g., ``Could you please specify for which instance I should provide a counterfactual?''. This gives us more flexibility and makes the dialogue flow more natural.


\begin{table*}[ht!]
    \centering

    \resizebox{\textwidth}{!}{%
        \begin{tabular}{rr|ccc|ccc|ccc}

        \toprule
        Dataset & & \multicolumn{3}{c|}{\data{BoolQ}} & \multicolumn{3}{c|}{\data{OLID}} & \multicolumn{3}{c}{\data{DailyDialog}} \\
        Parsing model & Size & \textit{dev} & \textit{dev-gpt} & \textit{test} & \textit{dev} & \textit{dev-gpt} & \textit{test} & \textit{dev} & \textit{dev-gpt} & \textit{test} \\

        \midrule
        Nearest Neighbors & - 
            & 34.69 & 35 & 34.02
            & 33.67 & 35 & 30.26
            & 36.73 & 37 & 32.51 \\
        \lm{GPT-Neo} & 2.7B 
            & \textbf{73} & 70 & 72.54
            & 71 & 72 & 67.11
            & 70 & 66 & 70.44 \\ 
        \lm{FLAN-T5-base} & 250M 
            & 71 & 71 & 74.18
            & 63 & 66 & 66.67
            & 66 & 63 & 75.86 \\ 
        \lm{BERT+Adapter} & 110M 
            & 72.55 & \textbf{76.86} & \textbf{79.33}
            & \textbf{72.55} & \textbf{76.86} & \textbf{84.25}
            & \textbf{72.55} & \textbf{77.69} & \textbf{83.94}\\
        
        \bottomrule
        \end{tabular}
    }
    \caption{Exact match parsing accuracy (in \%) for the datasets and their three partitions (human-authored \textit{dev} development data, \textit{dev-gpt} data augmented via \lm{GPT-3.5}, \textit{test} set created from questions asked by participants of the user study). \lm{GPT-Neo} uses $k=20$ shots in the prompt.
    }
    \label{tab:accuracy}
\end{table*} 

\section{NLP Models}
\label{sec:nlpmodels}

We selected three use cases in NLP with \lm{BERT}-type Transformer models trained on standard datasets, all of which we offer users to explore.

\subsection{Dialogue Act classification}

\data{DailyDialog} \cite{li-2017-dailydialog} is a multi-turn dialogue dataset that covers different topics related to our daily life (e.g., shopping, discussing vacation trips etc.). All conversations are human-written and there are 13,118 dialogues in total with 8 turns per dialogue on average. We limit the training set to the first 1,000 dialogues, the development set to 100 and the test set to 300 dialogues.

The dialogue act labels annotated in the dataset are as follows: Inform, Question, Directive and Commissive (see Figure \ref{fig:da-distr} for the distribution of labels). Inform is about providing information in the form of statements or questions. Question is used when the speaker wants to know something and actively asks for information. Directives are about requests, instructions, suggestions and acceptance or rejection of offers. Commissives are labeled when the speaker accepts or rejects requests or suggestions \cite{li-2017-dailydialog}. The Transformer model trained on \data{DailyDialog} achieves F1 score 68.7\% on the test set after 5 epochs of training with 5e-6 learning rate.

\subsection{Question answering}

We choose \data{BoolQ} \cite{clark-2019-boolq} as the representative dataset which has been analyzed in the explainability context in many works \cite[i.a.]{deyoung-2020-eraser,atanasova-2020-diagnostic,pezeshkpour-etal-2022-combining}. 
Each of the 16k examples consists of a question, a paragraph from a Wikipedia article, the title of that article, and a ``yes''/``no'' answer. 

We let its validation set (3.2k instances)\footnote{The ground truth labels for the test set are not available.} be predicted by a fine-tuned \lm{DistilBERT} \cite{sanh-2020-distilbert} model\footnote{\url{https://huggingface.co/andi611/distilbert-base-uncased-qa-boolq}} with an accuracy of 72.11\%. We choose a smaller model, because it is more easily deployable and more error-prone which increases the need for explanations.

\subsection{Hate speech detection}

Hate speech detection is a challenging task to determine user entries on social media if offensive. While better models for hate speech detection are continuously being developed, there is little research on the acceptability aspects of hate speech models. 
There have been a few studies on this task in the explainability literature, mostly using attributions or binary highlights \cite{mathew-2021-hatexplain,balkir-etal-2022-necessity,attanasio-etal-2022-benchmarking}.

\data{OLID} \cite{zampieri-2019-olid} is one of the common benchmark datasets and includes 14,100 tweets to be identified whether they are offensive.
Each row in \data{OLID} consists of text and label and the label indicates if the twitter text is ``offensive'' or ``non-offensive''. A fine-tuned \lm{mbert-olid-en}\footnote{\url{https://huggingface.co/sinhala-nlp/mbert-olid-en}} model is used to predict the validation set (2648 instances) and it can achieve an accuracy of 81.42\%.

\section{Interpretability and Analysis Components}
\label{sec:components}

For our implementation and experimental setup, we use the following tools and methods to realize the operations in Table~\ref{tab:operations}:

\paragraph{Attribution}
\citet{slack-2022-talktomodel} automatically select ``the most faithful feature importance method for users, unless a user specifically requests a certain technique''. We constrain feature importance to Integrated Gradients \cite{sundararajan-2017-axiomatic} saliency scores that we obtain from \sys{Captum} \cite{kokhlikyan-2020-captum}, which allows easy replacement with other saliency methods. 
The attributions are based on token-level as generated by the underlying model, e.g. \lm{BERT} in our experiments.
We also provide caching functionality to pre-compute and store the scores, thus reducing the inference time and mitigating expensive reruns on static inputs.

\paragraph{Perturbation}
For \textbf{counterfactual} generation, we use the official Hugging Face implementation of \sys{Polyjuice} \cite{wu-2021-polyjuice}\footnote{\url{https://huggingface.co/uw-hai/polyjuice}}.
\textbf{Adversarial examples} are generated via \sys{OpenAttack} \cite{zeng-2020-openattack}\footnote{\url{https://github.com/thunlp/OpenAttack}}, where we choose \sys{PWWS} \cite{ren-etal-2019-generating} as the attacker for our models on a single instance.
For \textbf{data augmentation} we use the \sys{NLPAug} library\footnote{\url{https://github.com/makcedward/nlpaug}} and replace some tokens in the text based on their embedding similarity computed with the \textit{bert-based-cased} model. The percentage of words that are augmented for each text is set to 0.3. 
We display the replaced words in bold, so that the user can easily distinguish between the original instance and the augmented one.

\paragraph{Rationalization}
As a baseline, we use the parsing model (\lm{GPTNeo}) in a \textit{zero-shot setup} to produce free-text explanations based on a concatenation of the input, the classification by the explained \lm{BERT}-type model \cite{marasovic-etal-2022-shot} and an instruction asking for an explanation. 
For an improved version, we produce plausible rationales from \lm{ChatGPT}\footnote{\url{https://platform.openai.com/docs/api-reference/chat}, March 23 version} and then prompt a \lm{Dolly-v2-3B}\footnote{\url{https://huggingface.co/databricks/dolly-v2-3b}} for \textit{few-shot} rationales. 
The rationales are pre-computed for all datasets.

\paragraph{Natural language understanding}
For computing the semantic \textbf{similarity}, we embed the data point using Sentence Transformers \cite{reimers-gurevych-2019-sentence} and compute the cosine similarity to other points (excluding the instance in question) in the respective dataset.
In order to retrieve frequent \textbf{keywords} from the whole dataset, we apply the stopwords set defined in \sys{NLTK} \cite{bird-2006-nltk} and get a word frequency set. The operation can then return the $n$ most frequent keywords, with $n$ being defined through the user query.

\begin{table}[t!]
    \centering

    \resizebox{\columnwidth}{!}{%
        \begin{tabular}{cr|cccc|}

        \toprule
        & Operations & \textbf{C}orr. & \textbf{H}elp. & \textbf{S}at. \\

        \midrule
        \multirow{5}{*}{\centering \rotatebox[origin=c]{90}{\textbf{Metadata}}}
        & Show example & 52.94 & 44.44 & 42.19 \\
        & Describe data & 89.66 & 87.27 & 87.72 \\
        & Count data & 56.41 & 44.44 & 45.83\\
        & True labels &58.82 & 64.71 & 72.22 \\
        & Model cards & 56.25 & 43.75 & 45.06 \\

        \midrule
        \multirow{5}{*}{\centering \rotatebox[origin=c]{90}{\textbf{Prediction}}}
        & Random prediction & 57.59 & 60.71 & 65.52 \\ 
        & Single/Dataset prediction & 53.42 & 53.52 & 54.17 \\
        & Likelihood & 62.86 & 67.50 & 63.41 \\
        & Performance & 72.50 & 65.79 & 76.19 \\
        & Mistakes & 81.25 & 68.75 & 77.09\\
        
        \midrule
        \multirow{2}{*}{\centering \rotatebox[origin=c]{90}{\textbf{NLU}}}
        & Similar examples & 53.57 & 45.61 & 62.50 \\
        & Keywords & 60.34 & 54.00 & 60.00\\
        
        \midrule
        \multirow{3}{*}{\centering \rotatebox[origin=c]{90}{\textbf{Expl.}}}
        & Feature importance & 55.88 & 42.25 & 50.00\\
        & Global feature importance & 50.00 & 50.00 & 31.32\\
        & Free-text rationale & 62.07 & 62.50 & 65.45\\

        \midrule
        \multirow{3}{*}{\centering \rotatebox[origin=c]{90}{\textbf{Pertb.}}}
        & Counterfactual & 40.00 & 27.03 & 21.62\\
        & Adversarial example & 61.90 & 40.00 & 37.50\\
        & Augmentation & 62.50 & 52.17 & 60.00\\
        
        \bottomrule
        \end{tabular}
    }
    \caption{Task A1 of the user study: Subjective ratings (\% positive) on correctness, helpfulness and satisfaction for single turns (responses in isolation), macro-averaged (each user has the same weight, regardless of how many ratings they gave). Custom input operations are averaged with their ``regular'' counterparts. 
    }
    \label{tab:singleratings}
\end{table}

\section{Evaluation}
\label{sec:eval}
We conduct our evaluation based on parsing accuracy and two user studies. After introducing the partitions we used to obtain the parsing (intent recognition) results (\S \ref{sec:autoeval}), we describe the setup of our human evaluation related to user experience and simulatability (\S \ref{sec:humeval}).

\subsection{Datasets}
\lm{FLAN-T5-base} and \lm{Adapter}-based models are trained on the \textit{train} set, which contains 505 pairs of user questions and prompts. We automatically extended the set for \lm{Adapter} by filling in all possible slots with the values from the datasets (Fig.~\ref{app:adapters-training-params}).
The \textit{train} set is a combination of manual creation by us and subsequent augmentation using \lm{ChatGPT}.
For evaluation, we created three more partitions (\textit{dev}, \textit{dev-gpt}, \textit{test}) to evaluate the parsing accuracy, as presented in Table \ref{tab:accuracy}. 
The \textit{dev} set has been manually created by us which consists of 102 pairs of user questions and parsed texts. To construct the \textit{dev-gpt} set, we leverage \lm{ChatGPT} to generate semantically similar examples extracted from \textit{dev} set. The \textit{test} set is obtained by collecting questions of participants who participated in the user study (\S \ref{sec:humeval}).
Unlike \ttm, our NLP datasets don't have a tabular format. Therefore, we had to adjust the parsing approach to be able to handle text inputs relevant to our NLP tasks.

\subsection{Automated evaluation: Intent recognition}
\label{sec:autoeval}
To answer the question of how well are user questions mapped onto the correct explanations and responses, for all three use cases, we compare the \lm{GPT-Neo-2.7B} parsing proposed in \citet{slack-2022-talktomodel} with our novel Adapter-based solution (\S \ref{sec:intent}) and also fine-tune a custom parsing model based on \lm{FLAN-T5-base} \cite{chung-2022-flan}.

\subsection{Human evaluation}
\label{sec:humeval}

\begin{table}[t!]
    \centering

        \begin{tabular}{r|cccc}

        \toprule
        Datasets & \textbf{C}orr. & \textbf{H}elp. & \textbf{S}at. & \textbf{F}lue. \\

        \midrule
        \data{BoolQ} & 3.6 & 3.3 & 2.5 & 3.1 \\
        \data{OLID} & 2.9 & 3.4 & 3.0 & 3.1 \\
        \data{DailyDialog} & 3.2 & 3.5 & 3.1 & 2.9 \\
        
        \bottomrule
        \end{tabular}
    \caption{Task A2 of the user study: Subjective ratings (Likert scale 1-5 with 1 being worst/disagree and 5 being best/fully agree) on correctness, helpfulness, satisfaction and fluency for entire dialogues.
    }
    \label{tab:dataratings}
\end{table}

Dialogue evaluation research has raised awareness of measuring flexibility and understanding among many other criteria. There exist automated metrics based on NLP models for assessing the quality of dialogues, but their correlation with human judgments needs to be improved on \cite{mehri-2022-nsf,siro-2022-user-satisfaction}.
While \ttm is focused on usability metrics (easiness, confidence, speed, likeliness to use), we target dialogue and explanation quality metrics.

\subsubsection{Subjective ratings}
\label{sec:evaluation-task-a}
A more precise way are user questionnaires \cite{kelly-2009-questionnaires}. We propose to focus on two types of questionnaires: Evaluating a user's experience \textbf{(1)} with one type of explanation (e.g. attribution), and \textbf{(2)} explanations in the context of the dialogue, with one type of downstream task (e.g., QA). An average of the second dimension will also provide a quality estimate for the overall system.

Concretely, we let 10 students with computational linguistics and computer science backgrounds\footnote{The participants of our user studies were recruited in-house: All of them were already working as research assistants in our institute and are compensated monthly based on national regulations. None of them had any prior experience with the explained models.} \textbf{explore} the tool and test out the available operations and then rate the following by giving a positive or negative review (\textbf{Task A}, App.~\ref{app:instructions-a}):
\begin{enumerate}[noitemsep,topsep=0pt]
    \item Correctness (C), helpfulness (H) and satisfaction (S) on the single-turn-level
    \item CHS and Fluency (F) on the dataset-level (when finishing the dialogue)
\end{enumerate}

\subsubsection{Simulatability}
\label{sec:evaluation-task-b}
We also conduct a simulatability evaluation (\textbf{Task B}, App.~\ref{app:instructions-b}), i.e. based on seeing an explanation and the original model input for a previously unseen instance. If a participant can correctly guess what the model predicted for that particular instance (which can also be a wrong classification) \cite{kim-2016-examples}, the explanation they saw would be deemed helpful. 
We can then express an objective quality estimate of each type of explanation in terms of simulation accuracy, both in isolation and in combination with other explanations.

Each participant (four authors of this paper + two students from Task A) received nine randomly chosen IDs (three from each dataset). The list of operations (Table~\ref{tab:sim}) is randomized for each ID, serving as the itinerary. After each response, the participant can decide to either perform the simulation (take the guess) or continue with the next in the list. After deciding on a simulated label, they are tasked to assign one helpfulness rating to each operation: 1~=~helpful; -1~=~not helpful; 0~=~unused.
Let $R$ be the set of all ratings $r_i \neq 0$ and $\mathbf{1}_{t}(x)$ our indicator function.
We then calculate our Helpfulness Ratio as follows:
\begin{center}
    Helpfulness Ratio = $\sum_{r \in R} \frac{\mathbf{1}_{1}(r)}{|R|}$.
\end{center}
Let $\hat{y}_i$ be the model prediction at index $i$ and $\tilde{y}_i$ the user's guess on the model prediction, then the simulation accuracy is
\begin{center}
    Sim(all) = $\sum_{i=1}^{|R|}\frac{ \mathbf{1}_{\hat{y}_i}(\tilde{y})}{|R|}$.
\end{center}
Filtering for all cases where the operation was deemed helpful:
\begin{center}
    Sim($t=1$) = $\sum_{i=1}^{|R|}\frac{\mathbf{1}_{\hat{y}_i}(\tilde{y}_i) \cdot \mathbf{1}_t(r_i)}{\mathbf{1}_t(r_i)}$.
\end{center}

\begin{table}[t!]
    \centering

    \resizebox{\columnwidth}{!}{%
        \begin{tabular}{r|cccc}

        \toprule
        \multirow{2}{*}{Explanation types} & \textbf{Sim} & \textbf{Sim} & \textbf{Help} & \#\textbf{Turns} \\
        & (all) & ($t=1$) & Ratio & Avg. \\

        \midrule
        Local feature importance & 91.43 & 93.10 & \textbf{82.86} & 3.85 \\
        Sent. feature importance & 90.00 & 94.44 & 60.00 & 3.84 \\
        Free-text rationale & \textbf{94.74} & \textbf{100.00} & 68.42 & \textbf{3.70} \\
        Counterfactual & 85.00 & 80.00 & 25.00 & 4.14 \\
        Adversarial example & 84.00 & 85.71 & 56.00 & 4.00 \\
        Similar examples & 88.46 & 87.50 & 61.54 & 4.00 \\
        
        \bottomrule
        \end{tabular}
    }
    \caption{Task B of the user study: Simulatability. Simulation accuracy (in \%), simulation accuracy for explanations deemed helpful (in \%), helpfulness ratio (in \%), average number of turns needed to make a decision.
    }
    \label{tab:sim}
\end{table}


\section{Results and discussion}

\paragraph{Parsing accuracy}
Table~\ref{tab:accuracy} shows that our Adapter-based approach (slot tagging and intent recognition) is able to outperform both the \lm{GPT-Neo} baseline and the fine-tuned \lm{FLAN-T5} models, 
using much fewer parameters and trained on the automatically augmented prompts with replaced slot values. 

\paragraph{Human preferences}
Table~\ref{tab:singleratings} reveals that most operations were positively received, but there are large differences between the subjective ratings of operations across all three aspects (CHS). We find that data description, performance and mistakes operations consistently perform highly, indicating that they're essential to model understanding.
Among the repertoire of explanation operations, free-text rationale scores highest on average, followed by augmentation and adversarial examples, while counterfactuals are at the bottom of the list. The \sys{Polyjuice} \lm{GPT} was often not able to come up with a perturbation (flipping the label) at all and we see the largest potential of improvement in the choice for a counterfactual generator.
The dialogue evaluation in Table~\ref{tab:dataratings} also solidifies the overall positive impressions. While \data{BoolQ} scored highest on Correctness, \data{DailyDialog} was the most favored in Helpfulness and Satisfaction. Fluency showed no differences, mostly because the generated texts are task-agnostic. 
Satisfaction was lowest across the three use cases. 
Although the operations were found to be helpful and correct, the satisfaction still leaves some room for improvements, likely due to high affordances (too much information at once) or low comprehensiveness. A more fine-grained evaluation \cite{siro-2022-user-satisfaction} might reveal whether this can be attributed to presentation mode, explanation quality or erroneous parses.

\paragraph{Simulatability}
Based on Table~\ref{tab:sim}, we can observe that the results align with the conclusions drawn from Table~\ref{tab:singleratings}. Specifically, free-text rationales provide the most assistance to users, while feature importance was a more useful operation for multi-turn simulation, compared to single-turn helpfulness ratings. 
On the other hand, counterfactual and adversarial examples are found to be least helpful, supporting the findings of Task A. Thus, their results may not consistently satisfy users' expectations.
We detected very few cases where one operation was sufficient. Combinations of explanations are essential: While attribution and rationales are needed to let users form their hypotheses about the model's behavior, counterfactuals and adversarial examples can be sanity checks that support or counter them \cite{hohman-2019-gamut}.
With $\text{Sim}(t=1)$, we detected that in some cases the explanations induced false trust and led the users to predict a different model output.


\subsection{Dataset with our results}

We compile a dataset from (1) our templates, (2) the automatically generated explanations, and (3) human feedback on the rationales presented through the interface. The research community can use these to perform further analyses and train more robust and human-aligned models.
We collected 1449 dialogue turns from feedback files (\hyperref[sec:evaluation-task-a]{Task A}) and 188 turns from the simulatability study (\hyperref[sec:evaluation-task-b]{Task B}). We provide a breakdown in App.~\ref{app:dataset}.


\section{Related Work}
\label{sec:rw}

\paragraph{Dialogue systems for interpretability in ML}

Table~\ref{tab:systems} shows the range of existing natural language interfaces and conversational agents for explanations.
Most notably, \sys{ConvXAI} \cite{shen-2023-convxai} very recently presented the first dialogue-based interpretability tool in the NLP domain. Their focus, however, is on the single task of LLMs as writing assistants. They also don't offer dataset exploration methods, their system is constrained to a single dataset (CODA-19) and they have not considered free-text rationalization, which we find is one of the most preferred types of operations.
\citet{dalvi-mishra-2022-teachme} proposed an interactive system to provide faithful explanations using previous interactions as a feedback. Despite being interactive, it does not provide feasibility of generating rationales on multiple queries subsequently.
\citet{bertrand-2023-selective} wrote a survey on prior studies on ``dialogic XAI'', 
while Fig. 6 of \citet{jacovi-2023-diagnosing} highlights that interactive interrogation is needed to construct complete explanation narratives: Feature attribution and counterfactuals complement each other, s.t. the users can build a generalizable mental model.

\paragraph{Visual interfaces for interpretability in NLP}
\sys{LIT} \cite{tenney-2020-lit}, \sys{Azimuth} \cite{gauthier-melancon-2022-azimuth}, \sys{IFAN} \cite{mosca-2023-ifan} and \sys{WebSHAP} \cite{wang-chau-2023-webshap} offer a broad range of explanations and interactive analyses on both local and global levels.
\sys{Robustness Gym} \cite{goel-etal-2021-robustness}, \sys{SEAL} \cite{rajani-2022-seal}, \sys{Evaluate} \cite{von-werra-2022-evaluate}, \sys{Interactive Model Cards} \cite{crisan-2022-interactive-model-cards} and \sys{DataLab} \cite{xiao-etal-2022-datalab} offer model evaluation, dataset analysis and accompanying visualization tools in practice.
There are overlaps with \ilang in the methods they integrate, but none of them offer a conversational interface like ours.

\paragraph{User studies on NLP interpretability}
Most influential to our study design are simulatability evaluations \cite{hase-bansal-2020-evaluating,nguyen-2018-comparing,gonzalez-2021-interaction,arora-2022-explain-edit-understand,das-2022-prototex,feldhus-2023-smv}.
In terms of preference ratings, \citet{strout-2019-human-rationales} evaluated how extractive rationales (discretized attributions) from different models are rated by human annotators.
Helpfulness and satisfaction ratings were used in \citet{schuff-2020-f1} and \citet{ray-2019-can-you-explain-that}.


\section{Conclusion}

We introduce our system, \ilang, which is a user-centered dialogue-based system for exploring the NLP datasets and model behavior. This system enables users to engage in multi-turn dialogues. Based on the findings from our conducted user study, we have determined that one-off explanations alone are usually not sufficient or beneficial. In many cases, users may require multiple explanations to obtain accurate predictions and gain a better understanding of the system's output.

Future work includes making the bot more pro-active, so that it can suggest new operations related to the user queries.
We also want to investigate the feasibility of using a singular LLM for all tasks (parsing, prediction, explanation generation\footnote{Operations have to be adapted in some cases, e.g., generating matrices for feature attribution \cite{sarti-2023-inseq} and counterfactuals without an external library \cite{chen-2023-disco}.}, response generation) over the modular setup that we currently employ;
Redesigning operations as API endpoints and training LLMs to call them \cite{lu-2023-chameleon,schick-2023-toolformer}, s.t. they can autonomously take care of the entire dialogue management at once.
Lastly, refining language models (increasing faithfulness or robustness, aligning with user expectations) through dialogues has gained traction \cite{lee-2023-xmd,madaan-2023-self-refine}. While we are already collecting valuable data, our framework misses an automated feedback loop to iteratively improve the models.


\section*{Limitations}
\ilang does not exhaust all interpretability methods, because understanding and integrating them requires a lot of resources. We see feature interactions, measurements of biases and component analysis as the most promising future work.

\ilang does not allow direct model comparison. The models are constrained to their datasets and the use cases are intended to be explored separately.

Users can enter custom inputs to get predicted and explained, but they can not modify the dataset on-the-fly, e.g., adding generated adversarial examples or augmentations directly to the current dataset and saving the updated version.

We do not offer a solution to mitigate biases or potential harmful effects of language models, but \ilang with its range of explanations is intended to point users into directions where the training data or model behavior is counter-intuitive. 

We use \lm{ChatGPT} only for (1) producing high-quality rationales to use in demonstrations (\S \ref{sec:components}) and (2) augmenting our intent recognition training data containing utterance-parse pairs (\S \ref{sec:intent}). We argue that these are legitimate use cases of \lm{ChatGPT}.
For almost every other part of \ilang, \lm{ChatGPT} is not applicable, though (see Footnote 1).
\ilang is a modular system and one of our goals is to have all modules be sourced from readily available tools. \lm{ChatGPT} can easily be swapped with a sufficiently strong rationalizer and data augmenter, as soon as they become available open source. At the time of implementing \ilang, however, we found that there is a large qualitative gap between \lm{ChatGPT} and open-source LLMs (\lm{Dolly}, \lm{GPT-Neo}) and that's why we opted to include it in these two parts of our framework.

\section*{Ethics Statement}

We incorporate \data{OLID} as one of our datasets, which may contain hateful or offensive words. However, it is important to note that we do not generate any new content that is hateful or offensive. Our usage of the \data{OLID} dataset is solely for the purpose of assessing the integration of the hate speech detection task to our system and generating plausible and useful explanations.

\section*{Acknowledgments}
We are indebted to Gokul Srinivasagan, Maximilian Dustin Nasert, Ammer Ayach, Christopher Ebert, Urs Alexander Peter, David Meier, Jo\~{a}o Lucas Mendes de Lemos Lins, Tim Patzelt, Elif Kara and Natalia Skachkova for their invaluable work as annotators.
We thank Leonhard Hennig, Malte Ostendorff, Jo\~{a}o Lucas Mendes de Lemos Lins and Maximilian Dustin Nasert for their review of earlier drafts and the reviewers of EMNLP 2023 for their helpful and rigorous feedback.
This work has been supported by the German Federal Ministry of Education and Research as part of the projects XAINES (01IW20005) and CORA4NLP (01IW20010) and the European Union as part of the AviaTor project (SEP-210730802).

\bibliography{anthology,custom}

\begin{thebibliography}{76}
\expandafter\ifx\csname natexlab\endcsname\relax\def\natexlab#1{#1}\fi

\bibitem[{Arora et~al.(2022)Arora, Pruthi, Sadeh, Cohen, Lipton, and
  Neubig}]{arora-2022-explain-edit-understand}
Siddhant Arora, Danish Pruthi, Norman Sadeh, William~W. Cohen, Zachary~C.
  Lipton, and Graham Neubig. 2022.
\newblock \href {https://doi.org/10.1609/aaai.v36i5.20464} {Explain, edit, and
  understand: Rethinking user study design for evaluating model explanations}.
\newblock In \emph{Proceedings of the AAAI Conference on Artificial
  Intelligence}, volume~36, pages 5277--5285.

\bibitem[{Atanasova et~al.(2020)Atanasova, Simonsen, Lioma, and
  Augenstein}]{atanasova-2020-diagnostic}
Pepa Atanasova, Jakob~Grue Simonsen, Christina Lioma, and Isabelle Augenstein.
  2020.
\newblock \href {https://doi.org/10.18653/v1/2020.emnlp-main.263} {A diagnostic
  study of explainability techniques for text classification}.
\newblock In \emph{Proceedings of the 2020 Conference on Empirical Methods in
  Natural Language Processing (EMNLP)}, pages 3256--3274, Online. Association
  for Computational Linguistics.

\bibitem[{Attanasio et~al.(2022)Attanasio, Nozza, Pastor, and
  Hovy}]{attanasio-etal-2022-benchmarking}
Giuseppe Attanasio, Debora Nozza, Eliana Pastor, and Dirk Hovy. 2022.
\newblock \href {https://doi.org/10.18653/v1/2022.nlppower-1.11} {Benchmarking
  post-hoc interpretability approaches for transformer-based misogyny
  detection}.
\newblock In \emph{Proceedings of NLP Power! The First Workshop on Efficient
  Benchmarking in NLP}, pages 100--112, Dublin, Ireland. Association for
  Computational Linguistics.

\bibitem[{Balkir et~al.(2022)Balkir, Nejadgholi, Fraser, and
  Kiritchenko}]{balkir-etal-2022-necessity}
Esma Balkir, Isar Nejadgholi, Kathleen Fraser, and Svetlana Kiritchenko. 2022.
\newblock \href {https://doi.org/10.18653/v1/2022.naacl-main.192} {Necessity
  and sufficiency for explaining text classifiers: A case study in hate speech
  detection}.
\newblock In \emph{Proceedings of the 2022 Conference of the North American
  Chapter of the Association for Computational Linguistics: Human Language
  Technologies}, pages 2672--2686, Seattle, United States. Association for
  Computational Linguistics.

\bibitem[{Bertrand et~al.(2023)Bertrand, Viard, Belloum, Eagan, and
  Maxwell}]{bertrand-2023-selective}
Astrid Bertrand, Tiphaine Viard, Rafik Belloum, James~R. Eagan, and Winston
  Maxwell. 2023.
\newblock \href {https://doi.org/10.1145/3544548.3581314} {On selective,
  mutable and dialogic {XAI}: A review of what users say about different types
  of interactive explanations}.
\newblock In \emph{Proceedings of the 2023 CHI Conference on Human Factors in
  Computing Systems}, CHI '23, New York, NY, USA. Association for Computing
  Machinery.

\bibitem[{Bird(2006)}]{bird-2006-nltk}
Steven Bird. 2006.
\newblock \href {https://doi.org/10.3115/1225403.1225421} {{NLTK}: The
  {N}atural {L}anguage {T}oolkit}.
\newblock In \emph{Proceedings of the {COLING}/{ACL} 2006 Interactive
  Presentation Sessions}, pages 69--72, Sydney, Australia. Association for
  Computational Linguistics.

\bibitem[{Camburu et~al.(2018)Camburu, Rockt\"{a}schel, Lukasiewicz, and
  Blunsom}]{camburu-2018-esnli}
Oana-Maria Camburu, Tim Rockt\"{a}schel, Thomas Lukasiewicz, and Phil Blunsom.
  2018.
\newblock \href
  {https://proceedings.neurips.cc/paper/2018/file/4c7a167bb329bd92580a99ce422d6fa6-Paper.pdf}
  {e-{SNLI}: Natural language inference with natural language explanations}.
\newblock In \emph{Advances in Neural Information Processing Systems},
  volume~31. Curran Associates, Inc.

\bibitem[{Chen et~al.(2023)Chen, Gao, Bosselut, Sabharwal, and
  Richardson}]{chen-2023-disco}
Zeming Chen, Qiyue Gao, Antoine Bosselut, Ashish Sabharwal, and Kyle
  Richardson. 2023.
\newblock \href {https://doi.org/10.18653/v1/2023.acl-long.302} {{DISCO}:
  Distilling counterfactuals with large language models}.
\newblock In \emph{Proceedings of the 61st Annual Meeting of the Association
  for Computational Linguistics (Volume 1: Long Papers)}, pages 5514--5528,
  Toronto, Canada. Association for Computational Linguistics.

\bibitem[{Chung et~al.(2022)Chung, Hou, Longpre, Zoph, Tay, Fedus, Li, Wang,
  Dehghani, Brahma, Webson, Gu, Dai, Suzgun, Chen, Chowdhery, Castro-Ros,
  Pellat, Robinson, Valter, Narang, Mishra, Yu, Zhao, Huang, Dai, Yu, Petrov,
  Chi, Dean, Devlin, Roberts, Zhou, Le, and Wei}]{chung-2022-flan}
Hyung~Won Chung, Le~Hou, Shayne Longpre, Barret Zoph, Yi~Tay, William Fedus,
  Yunxuan Li, Xuezhi Wang, Mostafa Dehghani, Siddhartha Brahma, Albert Webson,
  Shixiang~Shane Gu, Zhuyun Dai, Mirac Suzgun, Xinyun Chen, Aakanksha
  Chowdhery, Alex Castro-Ros, Marie Pellat, Kevin Robinson, Dasha Valter,
  Sharan Narang, Gaurav Mishra, Adams Yu, Vincent Zhao, Yanping Huang, Andrew
  Dai, Hongkun Yu, Slav Petrov, Ed~H. Chi, Jeff Dean, Jacob Devlin, Adam
  Roberts, Denny Zhou, Quoc~V. Le, and Jason Wei. 2022.
\newblock \href {https://arxiv.org/abs/2210.11416} {Scaling
  instruction-finetuned language models}.
\newblock \emph{arXiv}, abs/2210.11416.

\bibitem[{Clark et~al.(2019)Clark, Lee, Chang, Kwiatkowski, Collins, and
  Toutanova}]{clark-2019-boolq}
Christopher Clark, Kenton Lee, Ming-Wei Chang, Tom Kwiatkowski, Michael
  Collins, and Kristina Toutanova. 2019.
\newblock \href {https://doi.org/10.18653/v1/N19-1300} {{B}ool{Q}: Exploring
  the surprising difficulty of natural yes/no questions}.
\newblock In \emph{Proceedings of the 2019 Conference of the North {A}merican
  Chapter of the Association for Computational Linguistics: Human Language
  Technologies, Volume 1 (Long and Short Papers)}, pages 2924--2936,
  Minneapolis, Minnesota. Association for Computational Linguistics.

\bibitem[{Crisan et~al.(2022)Crisan, Drouhard, Vig, and
  Rajani}]{crisan-2022-interactive-model-cards}
Anamaria Crisan, Margaret Drouhard, Jesse Vig, and Nazneen Rajani. 2022.
\newblock \href {https://doi.org/10.1145/3531146.3533108} {Interactive model
  cards: A human-centered approach to model documentation}.
\newblock In \emph{2022 ACM Conference on Fairness, Accountability, and
  Transparency}, FAccT '22, page 427–439, New York, NY, USA. Association for
  Computing Machinery.

\bibitem[{Dalvi~Mishra et~al.(2022)Dalvi~Mishra, Tafjord, and
  Clark}]{dalvi-mishra-2022-teachme}
Bhavana Dalvi~Mishra, Oyvind Tafjord, and Peter Clark. 2022.
\newblock \href {https://aclanthology.org/2022.emnlp-main.644} {Towards
  teachable reasoning systems: Using a dynamic memory of user feedback for
  continual system improvement}.
\newblock In \emph{Proceedings of the 2022 Conference on Empirical Methods in
  Natural Language Processing}, pages 9465--9480, Abu Dhabi, United Arab
  Emirates. Association for Computational Linguistics.

\bibitem[{Das et~al.(2022)Das, Gupta, Kovatchev, Lease, and
  Li}]{das-2022-prototex}
Anubrata Das, Chitrank Gupta, Venelin Kovatchev, Matthew Lease, and Junyi~Jessy
  Li. 2022.
\newblock \href {https://doi.org/10.18653/v1/2022.acl-long.213} {{P}roto{TE}x:
  Explaining model decisions with prototype tensors}.
\newblock In \emph{Proceedings of the 60th Annual Meeting of the Association
  for Computational Linguistics (Volume 1: Long Papers)}, pages 2986--2997,
  Dublin, Ireland. Association for Computational Linguistics.

\bibitem[{DeYoung et~al.(2020)DeYoung, Jain, Rajani, Lehman, Xiong, Socher, and
  Wallace}]{deyoung-2020-eraser}
Jay DeYoung, Sarthak Jain, Nazneen~Fatema Rajani, Eric Lehman, Caiming Xiong,
  Richard Socher, and Byron~C. Wallace. 2020.
\newblock \href {https://doi.org/10.18653/v1/2020.acl-main.408} {{ERASER}: {A}
  benchmark to evaluate rationalized {NLP} models}.
\newblock In \emph{Proceedings of the 58th Annual Meeting of the Association
  for Computational Linguistics}, pages 4443--4458, Online. Association for
  Computational Linguistics.

\bibitem[{Dziri et~al.(2022)Dziri, Milton, Yu, Zaiane, and
  Reddy}]{dziri-2022-origin-hallucinations}
Nouha Dziri, Sivan Milton, Mo~Yu, Osmar Zaiane, and Siva Reddy. 2022.
\newblock \href {https://doi.org/10.18653/v1/2022.naacl-main.387} {On the
  origin of hallucinations in conversational models: Is it the datasets or the
  models?}
\newblock In \emph{Proceedings of the 2022 Conference of the North American
  Chapter of the Association for Computational Linguistics: Human Language
  Technologies}, pages 5271--5285, Seattle, United States. Association for
  Computational Linguistics.

\bibitem[{Ebrahimi et~al.(2018)Ebrahimi, Rao, Lowd, and
  Dou}]{ebrahimi-2018-hotflip}
Javid Ebrahimi, Anyi Rao, Daniel Lowd, and Dejing Dou. 2018.
\newblock \href {https://doi.org/10.18653/v1/P18-2006} {{H}ot{F}lip: White-box
  adversarial examples for text classification}.
\newblock In \emph{Proceedings of the 56th Annual Meeting of the Association
  for Computational Linguistics (Volume 2: Short Papers)}, pages 31--36,
  Melbourne, Australia. Association for Computational Linguistics.

\bibitem[{Feldhus et~al.(2023)Feldhus, Hennig, Nasert, Ebert, Schwarzenberg,
  and M{\"o}ller}]{feldhus-2023-smv}
Nils Feldhus, Leonhard Hennig, Maximilian~Dustin Nasert, Christopher Ebert,
  Robert Schwarzenberg, and Sebastian M{\"o}ller. 2023.
\newblock \href {https://aclanthology.org/2023.nlrse-1.4} {Saliency map
  verbalization: Comparing feature importance representations from model-free
  and instruction-based methods}.
\newblock In \emph{Proceedings of the 1st Workshop on Natural Language
  Reasoning and Structured Explanations (NLRSE)}, pages 30--46, Toronto,
  Canada. Association for Computational Linguistics.

\bibitem[{Feldhus et~al.(2022)Feldhus, Ravichandran, and
  Möller}]{feldhus-2022-mediators}
Nils Feldhus, Ajay~Madhavan Ravichandran, and Sebastian Möller. 2022.
\newblock \href {https://arxiv.org/abs/2206.06029} {Mediators: Conversational
  agents explaining {NLP} model behavior}.
\newblock In \emph{IJCAI 2022 - Workshop on Explainable Artificial Intelligence
  (XAI), Vienna, Austria}. International Joint Conferences on Artificial
  Intelligence Organization.

\bibitem[{Gauthier-Melan\c{c}on et~al.(2022)Gauthier-Melan\c{c}on,
  Marquez~Ayala, Brin, Tyler, Branchaud-Charron, Marinier, Grande, and
  Le}]{gauthier-melancon-2022-azimuth}
Gabrielle Gauthier-Melan\c{c}on, Orlando Marquez~Ayala, Lindsay Brin, Chris
  Tyler, Fr\'{e}d\'{e}ric Branchaud-Charron, Joseph Marinier, Karine Grande,
  and Di~Le. 2022.
\newblock \href {https://aclanthology.org/2022.emnlp-demos.30} {Azimuth:
  Systematic error analysis for text classification}.
\newblock In \emph{Proceedings of the 2022 Conference on Empirical Methods in
  Natural Language Processing: System Demonstrations}, pages 298--310, Abu
  Dhabi, UAE. Association for Computational Linguistics.

\bibitem[{Gebru et~al.(2021)Gebru, Morgenstern, Vecchione, Vaughan, Wallach,
  III, and Crawford}]{gebru-2021-datasheets}
Timnit Gebru, Jamie Morgenstern, Briana Vecchione, Jennifer~Wortman Vaughan,
  Hanna Wallach, Hal~Daum\'{e} III, and Kate Crawford. 2021.
\newblock \href {https://doi.org/10.1145/3458723} {Datasheets for datasets}.
\newblock \emph{Commun. ACM}, 64(12):86–92.

\bibitem[{Goel et~al.(2021)Goel, Rajani, Vig, Taschdjian, Bansal, and
  R{\'e}}]{goel-etal-2021-robustness}
Karan Goel, Nazneen~Fatema Rajani, Jesse Vig, Zachary Taschdjian, Mohit Bansal,
  and Christopher R{\'e}. 2021.
\newblock \href {https://doi.org/10.18653/v1/2021.naacl-demos.6} {Robustness
  gym: Unifying the {NLP} evaluation landscape}.
\newblock In \emph{Proceedings of the 2021 Conference of the North American
  Chapter of the Association for Computational Linguistics: Human Language
  Technologies: Demonstrations}, pages 42--55, Online. Association for
  Computational Linguistics.

\bibitem[{Gonz{\'a}lez et~al.(2021)Gonz{\'a}lez, Rogers, and
  S{\o}gaard}]{gonzalez-2021-interaction}
Ana~Valeria Gonz{\'a}lez, Anna Rogers, and Anders S{\o}gaard. 2021.
\newblock \href {https://doi.org/10.18653/v1/2021.findings-acl.259} {On the
  interaction of belief bias and explanations}.
\newblock In \emph{Findings of the Association for Computational Linguistics:
  ACL-IJCNLP 2021}, pages 2930--2942, Online. Association for Computational
  Linguistics.

\bibitem[{Hartmann et~al.(2022)Hartmann, Du, Feldhus, Kruijff-Korbayov{\'a},
  and Sonntag}]{hartmann-2022-xaines}
Mareike Hartmann, Han Du, Nils Feldhus, Ivana Kruijff-Korbayov{\'a}, and Daniel
  Sonntag. 2022.
\newblock \href {https://doi.org/10.1007/s13218-022-00780-8} {{XAINES}:
  Explaining {AI} with narratives}.
\newblock \emph{KI - K{\"u}nstliche Intelligenz}, 36(3):287--296.

\bibitem[{Hase and Bansal(2020)}]{hase-bansal-2020-evaluating}
Peter Hase and Mohit Bansal. 2020.
\newblock \href {https://doi.org/10.18653/v1/2020.acl-main.491} {Evaluating
  explainable {AI}: Which algorithmic explanations help users predict model
  behavior?}
\newblock In \emph{Proceedings of the 58th Annual Meeting of the Association
  for Computational Linguistics}, pages 5540--5552, Online. Association for
  Computational Linguistics.

\bibitem[{Hohman et~al.(2019)Hohman, Head, Caruana, DeLine, and
  Drucker}]{hohman-2019-gamut}
Fred Hohman, Andrew Head, Rich Caruana, Robert DeLine, and Steven~M. Drucker.
  2019.
\newblock \href {https://doi.org/10.1145/3290605.3300809} {Gamut: A design
  probe to understand how data scientists understand machine learning models}.
\newblock In \emph{Proceedings of the 2019 CHI Conference on Human Factors in
  Computing Systems}, CHI '19, page 1–13, New York, NY, USA. Association for
  Computing Machinery.

\bibitem[{Houlsby et~al.(2019)Houlsby, Giurgiu, Jastrzebski, Morrone,
  De~Laroussilhe, Gesmundo, Attariyan, and Gelly}]{houlsby-2019-parameter}
Neil Houlsby, Andrei Giurgiu, Stanislaw Jastrzebski, Bruna Morrone, Quentin
  De~Laroussilhe, Andrea Gesmundo, Mona Attariyan, and Sylvain Gelly. 2019.
\newblock \href {https://proceedings.mlr.press/v97/houlsby19a.html}
  {Parameter-efficient transfer learning for {NLP}}.
\newblock In \emph{Proceedings of the 36th International Conference on Machine
  Learning}, volume~97 of \emph{Proceedings of Machine Learning Research},
  pages 2790--2799. PMLR.

\bibitem[{Jacovi et~al.(2023)Jacovi, Bastings, Gehrmann, Goldberg, and
  Filippova}]{jacovi-2023-diagnosing}
Alon Jacovi, Jasmijn Bastings, Sebastian Gehrmann, Yoav Goldberg, and Katja
  Filippova. 2023.
\newblock \href {https://doi.org/10.1145/3593013.3593993} {Diagnosing ai
  explanation methods with folk concepts of behavior}.
\newblock In \emph{Proceedings of the 2023 ACM Conference on Fairness,
  Accountability, and Transparency}, FAccT '23, page 247, New York, NY, USA.
  Association for Computing Machinery.

\bibitem[{Kelly et~al.(2009)Kelly, Kantor, Morse, Scholtz, and
  Sun}]{kelly-2009-questionnaires}
Diane Kelly, Paul~B. Kantor, Emile~L. Morse, Jean Scholtz, and Ying Sun. 2009.
\newblock \href {https://doi.org/10.1017/S1351324908004932} {Questionnaires for
  eliciting evaluation data from users of interactive question answering
  systems}.
\newblock \emph{Natural Language Engineering}, 15(1):119–141.

\bibitem[{Kim et~al.(2016)Kim, Khanna, and Koyejo}]{kim-2016-examples}
Been Kim, Rajiv Khanna, and Oluwasanmi~O Koyejo. 2016.
\newblock \href
  {https://proceedings.neurips.cc/paper/2016/file/5680522b8e2bb01943234bce7bf84534-Paper.pdf}
  {Examples are not enough, learn to criticize! criticism for
  interpretability}.
\newblock In \emph{Advances in Neural Information Processing Systems},
  volume~29. Curran Associates, Inc.

\bibitem[{Koh and Liang(2017)}]{koh-liang-2017-influence}
Pang~Wei Koh and Percy Liang. 2017.
\newblock \href {https://proceedings.mlr.press/v70/koh17a.html} {Understanding
  black-box predictions via influence functions}.
\newblock In \emph{Proceedings of the 34th International Conference on Machine
  Learning}, volume~70 of \emph{Proceedings of Machine Learning Research},
  pages 1885--1894. PMLR.

\bibitem[{Ku{\'{z}}ba and Biecek(2020)}]{kuzba-biecek-2020-what-would}
Micha{\l} Ku{\'{z}}ba and Przemys{\l}aw Biecek. 2020.
\newblock \href {https://doi.org/10.1007/978-3-030-65965-3_30} {What would you
  ask the machine learning model? identification of user needs for model
  explanations based on human-model conversations}.
\newblock In \emph{ECML PKDD 2020 Workshops}, pages 447--459, Cham. Springer
  International Publishing.

\bibitem[{Lakkaraju et~al.(2022)Lakkaraju, Slack, Chen, Tan, and
  Singh}]{lakkaraju-2022-rethinking}
Himabindu Lakkaraju, Dylan Slack, Yuxin Chen, Chenhao Tan, and Sameer Singh.
  2022.
\newblock \href {https://arxiv.org/abs/2202.01875} {Rethinking explainability
  as a dialogue: {A} practitioner's perspective}.
\newblock \emph{HCAI @ NeurIPS 2022}.

\bibitem[{Lee et~al.(2023)Lee, Kadakia, Joshi, Chan, Liu, Narahari, Shibuya,
  Mitani, Sekiya, Pujara, and Ren}]{lee-2023-xmd}
Dong-Ho Lee, Akshen Kadakia, Brihi Joshi, Aaron Chan, Ziyi Liu, Kiran Narahari,
  Takashi Shibuya, Ryosuke Mitani, Toshiyuki Sekiya, Jay Pujara, and Xiang Ren.
  2023.
\newblock \href {https://arxiv.org/abs/2210.16978} {{XMD}: An end-to-end
  framework for interactive explanation-based debugging of {NLP} models}.
\newblock In \emph{Proceedings of the 61st Annual Meeting of the Association
  for Computational Linguistics: System Demonstrations}, Toronto, Canada.
  Association for Computational Linguistics.

\bibitem[{Li et~al.(2017)Li, Su, Shen, Li, Cao, and Niu}]{li-2017-dailydialog}
Yanran Li, Hui Su, Xiaoyu Shen, Wenjie Li, Ziqiang Cao, and Shuzi Niu. 2017.
\newblock \href {https://aclanthology.org/I17-1099} {{D}aily{D}ialog: A
  manually labelled multi-turn dialogue dataset}.
\newblock In \emph{Proceedings of the Eighth International Joint Conference on
  Natural Language Processing (Volume 1: Long Papers)}, pages 986--995, Taipei,
  Taiwan. Asian Federation of Natural Language Processing.

\bibitem[{Lu et~al.(2023)Lu, Peng, Cheng, Galley, Chang, Wu, Zhu, and
  Gao}]{lu-2023-chameleon}
Pan Lu, Baolin Peng, Hao Cheng, Michel Galley, Kai-Wei Chang, Ying~Nian Wu,
  Song-Chun Zhu, and Jianfeng Gao. 2023.
\newblock \href {https://arxiv.org/abs/2304.09842} {Chameleon: Plug-and-play
  compositional reasoning with large language models}.
\newblock \emph{arXiv}, abs/2304.09842.

\bibitem[{Madaan et~al.(2023)Madaan, Tandon, Gupta, Hallinan, Gao, Wiegreffe,
  Alon, Dziri, Prabhumoye, Yang, Welleck, Majumder, Gupta, Yazdanbakhsh, and
  Clark}]{madaan-2023-self-refine}
Aman Madaan, Niket Tandon, Prakhar Gupta, Skyler Hallinan, Luyu Gao, Sarah
  Wiegreffe, Uri Alon, Nouha Dziri, Shrimai Prabhumoye, Yiming Yang, Sean
  Welleck, Bodhisattwa~Prasad Majumder, Shashank Gupta, Amir Yazdanbakhsh, and
  Peter Clark. 2023.
\newblock \href {https://arxiv.org/abs/2303.17651} {Self-{R}efine: Iterative
  refinement with self-feedback}.
\newblock In \emph{Advances in Neural Information Processing Systems}.

\bibitem[{Madsen et~al.(2022)Madsen, Reddy, and Chandar}]{madsen-2022-post-hoc}
Andreas Madsen, Siva Reddy, and Sarath Chandar. 2022.
\newblock \href {https://doi.org/10.1145/3546577} {Post-hoc interpretability
  for neural {NLP}: A survey}.
\newblock \emph{ACM Comput. Surv.}

\bibitem[{Malandri et~al.(2022)Malandri, Mercorio, Mezzanzanica, and
  Nobani}]{malandri-2022-convxai}
Lorenzo Malandri, Fabio Mercorio, Mario Mezzanzanica, and Navid Nobani. 2022.
\newblock \href {https://doi.org/10.1007/s12559-022-10067-7} {Conv{XAI}: a
  system for multimodal interaction with any black-box explainer}.
\newblock \emph{Cognitive Computation}, 15(2):613--644.

\bibitem[{Marasovic et~al.(2022)Marasovic, Beltagy, Downey, and
  Peters}]{marasovic-etal-2022-shot}
Ana Marasovic, Iz~Beltagy, Doug Downey, and Matthew Peters. 2022.
\newblock \href {https://doi.org/10.18653/v1/2022.findings-naacl.31} {Few-shot
  self-rationalization with natural language prompts}.
\newblock In \emph{Findings of the Association for Computational Linguistics:
  NAACL 2022}, pages 410--424, Seattle, United States. Association for
  Computational Linguistics.

\bibitem[{Mathew et~al.(2021)Mathew, Saha, Yimam, Biemann, Goyal, and
  Mukherjee}]{mathew-2021-hatexplain}
Binny Mathew, Punyajoy Saha, Seid~Muhie Yimam, Chris Biemann, Pawan Goyal, and
  Animesh Mukherjee. 2021.
\newblock \href {https://ojs.aaai.org/index.php/AAAI/article/view/17745}
  {Hatexplain: A benchmark dataset for explainable hate speech detection}.
\newblock \emph{Proceedings of the AAAI Conference on Artificial Intelligence},
  35(17):14867--14875.

\bibitem[{Mehri et~al.(2022)Mehri, Choi, D’Haro, Deriu, Esk{\'e}nazi, Gasic,
  Georgila, Hakkani-T{\"u}r, Li, Rieser, Shaikh, Traum, Yeh, Yu, Zhang, and
  Zhang}]{mehri-2022-nsf}
Shikib Mehri, Jinho Choi, L.~F. D’Haro, Jan Deriu, Maxine Esk{\'e}nazi,
  Milica Gasic, Kallirroi Georgila, Dilek~Z. Hakkani-T{\"u}r, Zekang Li, Verena
  Rieser, Samira Shaikh, David~R. Traum, Yi-Ting Yeh, Zhou Yu, Yizhe Zhang, and
  Chen Zhang. 2022.
\newblock \href {https://doi.org/https://doi.org/10.48550/arXiv.2203.10012}
  {Report from the {NSF} future directions workshop on automatic evaluation of
  dialog: Research directions and challenges}.
\newblock \emph{arXiv}, abs/2203.10012.

\bibitem[{Miglani et~al.(2023)Miglani, Yang, Markosyan, Garcia-Olano, and
  Kokhlikyan}]{kokhlikyan-2020-captum}
Vivek Miglani, Aobo Yang, Aram~H. Markosyan, Diego Garcia-Olano, and Narine
  Kokhlikyan. 2023.
\newblock \href {https://openreview.net/forum?id=WBRpxtd3l4} {Using captum to
  explain generative language models}.
\newblock In \emph{Proceedings of the 3rd Workshop for Natural Language
  Processing Open Source Software (NLP-OSS)}, Singapore. Association for
  Computational Linguistics.

\bibitem[{Miller(2019)}]{miller-2019-explanation}
Tim Miller. 2019.
\newblock \href {https://doi.org/10.1016/j.artint.2018.07.007} {Explanation in
  artificial intelligence: Insights from the social sciences}.
\newblock \emph{Artificial Intelligence}, 267:1--38.

\bibitem[{Mitchell et~al.(2019)Mitchell, Wu, Zaldivar, Barnes, Vasserman,
  Hutchinson, Spitzer, Raji, and Gebru}]{mitchell-2019-model-cards}
Margaret Mitchell, Simone Wu, Andrew Zaldivar, Parker Barnes, Lucy Vasserman,
  Ben Hutchinson, Elena Spitzer, Inioluwa~Deborah Raji, and Timnit Gebru. 2019.
\newblock \href {https://doi.org/10.1145/3287560.3287596} {Model cards for
  model reporting}.
\newblock In \emph{Proceedings of the Conference on Fairness, Accountability,
  and Transparency}, FAT* '19, page 220–229, New York, NY, USA. Association
  for Computing Machinery.

\bibitem[{Mosca et~al.(2023)Mosca, Dementieva, Ajdari, Kummeth, Gringauz, and
  Groh}]{mosca-2023-ifan}
Edoardo Mosca, Daryna Dementieva, Tohid~Ebrahim Ajdari, Maximilian Kummeth,
  Kirill Gringauz, and Georg Groh. 2023.
\newblock \href {https://arxiv.org/abs/2303.03124} {{IFAN}: An
  explainability-focused interaction framework for humans and {NLP} models}.
\newblock In \emph{Proceedings of the 3rd Conference of the Asia-Pacific
  Chapter of the Association for Computational Linguistics and the 12th
  International Joint Conference on Natural Language Processing: System
  Demonstrations}, Bali, Indonesia. Association for Computational Linguistics.

\bibitem[{Nguyen(2018)}]{nguyen-2018-comparing}
Dong Nguyen. 2018.
\newblock \href {https://doi.org/10.18653/v1/N18-1097} {Comparing automatic and
  human evaluation of local explanations for text classification}.
\newblock In \emph{Proceedings of the 2018 Conference of the North {A}merican
  Chapter of the Association for Computational Linguistics: Human Language
  Technologies, Volume 1 (Long Papers)}, pages 1069--1078, New Orleans,
  Louisiana. Association for Computational Linguistics.

\bibitem[{Nguyen et~al.(2023)Nguyen, Schl{\"{o}}tterer, and
  Seifert}]{nguyen-2022-xagent}
Van~Bach Nguyen, J{\"{o}}rg Schl{\"{o}}tterer, and Christin Seifert. 2023.
\newblock \href {https://doi.org/10.48550/arXiv.2209.02552} {Explaining machine
  learning models in natural conversations: Towards a conversational {XAI}
  agent}.
\newblock In \emph{The World Conference on eXplainable Artificial Intelligence
  2023 (XAI-2023)}, Lisbon, Portugal.

\bibitem[{Pezeshkpour et~al.(2022)Pezeshkpour, Jain, Singh, and
  Wallace}]{pezeshkpour-etal-2022-combining}
Pouya Pezeshkpour, Sarthak Jain, Sameer Singh, and Byron Wallace. 2022.
\newblock \href {https://doi.org/10.18653/v1/2022.findings-acl.153} {Combining
  feature and instance attribution to detect artifacts}.
\newblock In \emph{Findings of the Association for Computational Linguistics:
  ACL 2022}, pages 1934--1946, Dublin, Ireland. Association for Computational
  Linguistics.

\bibitem[{Pfeiffer et~al.(2020)Pfeiffer, R{\"u}ckl{\'e}, Poth, Kamath,
  Vuli{\'c}, Ruder, Cho, and Gurevych}]{pfeiffer-2020-adapterhub}
Jonas Pfeiffer, Andreas R{\"u}ckl{\'e}, Clifton Poth, Aishwarya Kamath, Ivan
  Vuli{\'c}, Sebastian Ruder, Kyunghyun Cho, and Iryna Gurevych. 2020.
\newblock \href {https://doi.org/10.18653/v1/2020.emnlp-demos.7}
  {{A}dapter{H}ub: A framework for adapting transformers}.
\newblock In \emph{Proceedings of the 2020 Conference on Empirical Methods in
  Natural Language Processing: System Demonstrations}, pages 46--54, Online.
  Association for Computational Linguistics.

\bibitem[{Rajani et~al.(2022)Rajani, Liang, Chen, Mitchell, and
  Zou}]{rajani-2022-seal}
Nazneen Rajani, Weixin Liang, Lingjiao Chen, Margaret Mitchell, and James Zou.
  2022.
\newblock \href {https://aclanthology.org/2022.emnlp-demos.36} {{SEAL}:
  Interactive tool for systematic error analysis and labeling}.
\newblock In \emph{Proceedings of the 2022 Conference on Empirical Methods in
  Natural Language Processing: System Demonstrations}, pages 359--370, Abu
  Dhabi, UAE. Association for Computational Linguistics.

\bibitem[{Ray et~al.(2019)Ray, Yao, Kumar, Divakaran, and
  Burachas}]{ray-2019-can-you-explain-that}
Arijit Ray, Yi~Yao, Rakesh Kumar, Ajay Divakaran, and Giedrius Burachas. 2019.
\newblock \href {https://arxiv.org/abs/1904.03285} {Can you explain that? lucid
  explanations help human-ai collaborative image retrieval}.
\newblock In \emph{Proceedings of the AAAI Conference on Human Computation and
  Crowdsourcing}, volume~7, pages 153--161.

\bibitem[{Reimers and Gurevych(2019)}]{reimers-gurevych-2019-sentence}
Nils Reimers and Iryna Gurevych. 2019.
\newblock \href {https://doi.org/10.18653/v1/D19-1410} {Sentence-{BERT}:
  Sentence embeddings using {S}iamese {BERT}-networks}.
\newblock In \emph{Proceedings of the 2019 Conference on Empirical Methods in
  Natural Language Processing and the 9th International Joint Conference on
  Natural Language Processing (EMNLP-IJCNLP)}, pages 3982--3992, Hong Kong,
  China. Association for Computational Linguistics.

\bibitem[{Ren et~al.(2019)Ren, Deng, He, and Che}]{ren-etal-2019-generating}
Shuhuai Ren, Yihe Deng, Kun He, and Wanxiang Che. 2019.
\newblock \href {https://doi.org/10.18653/v1/P19-1103} {Generating natural
  language adversarial examples through probability weighted word saliency}.
\newblock In \emph{Proceedings of the 57th Annual Meeting of the Association
  for Computational Linguistics}, pages 1085--1097, Florence, Italy.
  Association for Computational Linguistics.

\bibitem[{R{\"o}nnqvist et~al.(2022)R{\"o}nnqvist, Kyr{\"o}l{\"a}inen, Myntti,
  Ginter, and Laippala}]{ronnqvist-etal-2022-explaining}
Samuel R{\"o}nnqvist, Aki-Juhani Kyr{\"o}l{\"a}inen, Amanda Myntti, Filip
  Ginter, and Veronika Laippala. 2022.
\newblock \href {https://doi.org/10.18653/v1/2022.findings-acl.85} {Explaining
  classes through stable word attributions}.
\newblock In \emph{Findings of the Association for Computational Linguistics:
  ACL 2022}, pages 1063--1074, Dublin, Ireland. Association for Computational
  Linguistics.

\bibitem[{Ross et~al.(2022)Ross, Wu, Peng, Peters, and
  Gardner}]{ross-etal-2022-tailor}
Alexis Ross, Tongshuang Wu, Hao Peng, Matthew Peters, and Matt Gardner. 2022.
\newblock \href {https://doi.org/10.18653/v1/2022.acl-long.228} {Tailor:
  Generating and perturbing text with semantic controls}.
\newblock In \emph{Proceedings of the 60th Annual Meeting of the Association
  for Computational Linguistics (Volume 1: Long Papers)}, pages 3194--3213,
  Dublin, Ireland. Association for Computational Linguistics.

\bibitem[{Sanh et~al.(2019)Sanh, Debut, Chaumond, and
  Wolf}]{sanh-2020-distilbert}
Victor Sanh, Lysandre Debut, Julien Chaumond, and Thomas Wolf. 2019.
\newblock \href {https://doi.org/10.48550/arXiv.1910.01108} {Distil{BERT}, a
  distilled version of {BERT}: smaller, faster, cheaper and lighter}.
\newblock In \emph{5th Workshop on Energy Efficient Machine Learning and
  Cognitive Computing - NeurIPS 2019}.

\bibitem[{Sarti et~al.(2023)Sarti, Feldhus, Sickert, and van~der
  Wal}]{sarti-2023-inseq}
Gabriele Sarti, Nils Feldhus, Ludwig Sickert, and Oskar van~der Wal. 2023.
\newblock \href {https://aclanthology.org/2023.acl-demo.40} {Inseq: An
  interpretability toolkit for sequence generation models}.
\newblock In \emph{Proceedings of the 61st Annual Meeting of the Association
  for Computational Linguistics (Volume 3: System Demonstrations)}, pages
  421--435, Toronto, Canada. Association for Computational Linguistics.

\bibitem[{Schick et~al.(2023)Schick, Dwivedi-Yu, Dessì, Raileanu, Lomeli,
  Zettlemoyer, Cancedda, and Scialom}]{schick-2023-toolformer}
Timo Schick, Jane Dwivedi-Yu, Roberto Dessì, Roberta Raileanu, Maria Lomeli,
  Luke Zettlemoyer, Nicola Cancedda, and Thomas Scialom. 2023.
\newblock \href {https://arxiv.org/abs/2302.04761} {Toolformer: Language models
  can teach themselves to use tools}.
\newblock \emph{arXiv}, abs/2302.04761.

\bibitem[{Schuff et~al.(2020)Schuff, Adel, and Vu}]{schuff-2020-f1}
Hendrik Schuff, Heike Adel, and Ngoc~Thang Vu. 2020.
\newblock \href {https://doi.org/10.18653/v1/2020.emnlp-main.575} {{F}1 is
  {N}ot {E}nough! {M}odels and {E}valuation {T}owards {U}ser-{C}entered
  {E}xplainable {Q}uestion {A}nswering}.
\newblock In \emph{Proceedings of the 2020 Conference on Empirical Methods in
  Natural Language Processing (EMNLP)}, pages 7076--7095, Online. Association
  for Computational Linguistics.

\bibitem[{Schuff et~al.(2022)Schuff, Jacovi, Adel, Goldberg, and
  Vu}]{schuff-2022-human}
Hendrik Schuff, Alon Jacovi, Heike Adel, Yoav Goldberg, and Ngoc~Thang Vu.
  2022.
\newblock \href {https://doi.org/10.1145/3531146.3533127} {Human interpretation
  of saliency-based explanation over text}.
\newblock In \emph{2022 ACM Conference on Fairness, Accountability, and
  Transparency}, FAccT '22, page 611–636, New York, NY, USA. Association for
  Computing Machinery.

\bibitem[{Shen et~al.(2023)Shen, Huang, Wu, and Huang}]{shen-2023-convxai}
Hua Shen, Chieh-Yang Huang, Tongshuang Wu, and Ting-Hao~Kenneth Huang. 2023.
\newblock \href {https://doi.org/10.1145/3584931.3607492} {Conv{XAI}:
  Delivering heterogeneous {AI} explanations via conversations to support
  human-{AI} scientific writing}.
\newblock In \emph{Computer Supported Cooperative Work and Social Computing},
  CSCW '23 Companion, page 384–387, New York, NY, USA. Association for
  Computing Machinery.

\bibitem[{Siro et~al.(2022)Siro, Aliannejadi, and
  de~Rijke}]{siro-2022-user-satisfaction}
Clemencia Siro, Mohammad Aliannejadi, and Maarten de~Rijke. 2022.
\newblock \href {https://doi.org/10.1145/3477495.3531798} {Understanding user
  satisfaction with task-oriented dialogue systems}.
\newblock In \emph{Proceedings of the 45th International ACM SIGIR Conference
  on Research and Development in Information Retrieval}, SIGIR '22, page
  2018–2023, New York, NY, USA. Association for Computing Machinery.

\bibitem[{Slack et~al.(2023)Slack, Krishna, Lakkaraju, and
  Singh}]{slack-2022-talktomodel}
Dylan Slack, Satyapriya Krishna, Himabindu Lakkaraju, and Sameer Singh. 2023.
\newblock \href {https://doi.org/10.1038/s42256-023-00692-8} {Explaining
  machine learning models with interactive natural language conversations using
  {T}alk{T}o{M}odel}.
\newblock \emph{Nature Machine Intelligence}.

\bibitem[{Strout et~al.(2019)Strout, Zhang, and
  Mooney}]{strout-2019-human-rationales}
Julia Strout, Ye~Zhang, and Raymond Mooney. 2019.
\newblock \href {https://doi.org/10.18653/v1/W19-4807} {Do human rationales
  improve machine explanations?}
\newblock In \emph{Proceedings of the 2019 ACL Workshop BlackboxNLP: Analyzing
  and Interpreting Neural Networks for NLP}, pages 56--62, Florence, Italy.
  Association for Computational Linguistics.

\bibitem[{Sundararajan et~al.(2017)Sundararajan, Taly, and
  Yan}]{sundararajan-2017-axiomatic}
Mukund Sundararajan, Ankur Taly, and Qiqi Yan. 2017.
\newblock \href {https://proceedings.mlr.press/v70/sundararajan17a.html}
  {Axiomatic attribution for deep networks}.
\newblock In \emph{Proceedings of the 34th International Conference on Machine
  Learning}, volume~70 of \emph{Proceedings of Machine Learning Research},
  pages 3319--3328. PMLR.

\bibitem[{Tenney et~al.(2020)Tenney, Wexler, Bastings, Bolukbasi, Coenen,
  Gehrmann, Jiang, Pushkarna, Radebaugh, Reif, and Yuan}]{tenney-2020-lit}
Ian Tenney, James Wexler, Jasmijn Bastings, Tolga Bolukbasi, Andy Coenen,
  Sebastian Gehrmann, Ellen Jiang, Mahima Pushkarna, Carey Radebaugh, Emily
  Reif, and Ann Yuan. 2020.
\newblock \href {https://doi.org/10.18653/v1/2020.emnlp-demos.15} {The language
  interpretability tool: Extensible, interactive visualizations and analysis
  for {NLP} models}.
\newblock In \emph{Proceedings of the 2020 Conference on Empirical Methods in
  Natural Language Processing: System Demonstrations}, pages 107--118, Online.
  Association for Computational Linguistics.

\bibitem[{Torri(2021)}]{torri-2021-textual-explanations}
Vittorio Torri. 2021.
\newblock \href {http://hdl.handle.net/10589/181513} {Textual e{X}planations
  for intuitive machine learning}.
\newblock Master's thesis, Politecnico di Milano, dec.

\bibitem[{von Werra et~al.(2022)von Werra, Tunstall, Thakur, Luccioni, Thrush,
  Piktus, Marty, Rajani, Mustar, Ngo, Sanseviero, Sasko, Villanova, Lhoest,
  Chaumond, Mitchell, Rush, Wolf, and Kiela}]{von-werra-2022-evaluate}
Leandro von Werra, Lewis Tunstall, Abhishek Thakur, Alexandra~Sasha Luccioni,
  Tristan Thrush, Aleksandra Piktus, Felix Marty, Nazneen Rajani, Victor
  Mustar, Helen Ngo, Omar Sanseviero, Mario Sasko, Albert Villanova, Quentin
  Lhoest, Julien Chaumond, Margaret Mitchell, Alexander~M. Rush, Thomas Wolf,
  and Douwe Kiela. 2022.
\newblock \href {https://arxiv.org/abs/2210.01970} {Evaluate \& evaluation on
  the hub: Better best practices for data and model measurement}.
\newblock In \emph{Proceedings of the 2022 Conference on Empirical Methods in
  Natural Language Processing: System Demonstrations}, pages 128--136, Abu
  Dhabi, UAE. Association for Computational Linguistics.

\bibitem[{Wang and Chau(2023)}]{wang-chau-2023-webshap}
Zijie~J. Wang and Duen~Horng Chau. 2023.
\newblock \href {https://doi.org/10.1145/3543873.3587362} {Webshap: Towards
  explaining any machine learning models anywhere}.
\newblock In \emph{Companion Proceedings of the ACM Web Conference 2023}, WWW
  '23 Companion, page 262–266, New York, NY, USA. Association for Computing
  Machinery.

\bibitem[{Weld and Bansal(2019)}]{weld-bansal-2019-crafting}
Daniel~S. Weld and Gagan Bansal. 2019.
\newblock \href {https://doi.org/10.1145/3282486} {The challenge of crafting
  intelligible intelligence}.
\newblock \emph{Commun. ACM}, 62(6):70–79.

\bibitem[{Werner(2020)}]{werner-2020-eric}
Christian Werner. 2020.
\newblock \href {http://ceur-ws.org/Vol-2578/ETMLP3.pdf} {Explainable ai
  through rule-based interactive conversation}.
\newblock In \emph{Proceedings of the Workshops of the EDBT/ICDT 2020 Joint
  Conference}.

\bibitem[{Wiegreffe et~al.(2022)Wiegreffe, Hessel, Swayamdipta, Riedl, and
  Choi}]{wiegreffe-etal-2022-reframing}
Sarah Wiegreffe, Jack Hessel, Swabha Swayamdipta, Mark Riedl, and Yejin Choi.
  2022.
\newblock \href {https://doi.org/10.18653/v1/2022.naacl-main.47} {Reframing
  human-{AI} collaboration for generating free-text explanations}.
\newblock In \emph{Proceedings of the 2022 Conference of the North American
  Chapter of the Association for Computational Linguistics: Human Language
  Technologies}, pages 632--658, Seattle, United States. Association for
  Computational Linguistics.

\bibitem[{Wu et~al.(2021)Wu, Ribeiro, Heer, and Weld}]{wu-2021-polyjuice}
Tongshuang Wu, Marco~Tulio Ribeiro, Jeffrey Heer, and Daniel Weld. 2021.
\newblock \href {https://doi.org/10.18653/v1/2021.acl-long.523} {Polyjuice:
  Generating counterfactuals for explaining, evaluating, and improving models}.
\newblock In \emph{Proceedings of the 59th Annual Meeting of the Association
  for Computational Linguistics and the 11th International Joint Conference on
  Natural Language Processing (Volume 1: Long Papers)}, pages 6707--6723,
  Online. Association for Computational Linguistics.

\bibitem[{Xiao et~al.(2022)Xiao, Fu, Yuan, Viswanathan, Liu, Liu, Neubig, and
  Liu}]{xiao-etal-2022-datalab}
Yang Xiao, Jinlan Fu, Weizhe Yuan, Vijay Viswanathan, Zhoumianze Liu, Yixin
  Liu, Graham Neubig, and Pengfei Liu. 2022.
\newblock \href {https://doi.org/10.18653/v1/2022.acl-demo.18} {{D}ata{L}ab: A
  platform for data analysis and intervention}.
\newblock In \emph{Proceedings of the 60th Annual Meeting of the Association
  for Computational Linguistics: System Demonstrations}, pages 182--195,
  Dublin, Ireland. Association for Computational Linguistics.

\bibitem[{Zampieri et~al.(2019)Zampieri, Malmasi, Nakov, Rosenthal, Farra, and
  Kumar}]{zampieri-2019-olid}
Marcos Zampieri, Shervin Malmasi, Preslav Nakov, Sara Rosenthal, Noura Farra,
  and Ritesh Kumar. 2019.
\newblock \href {https://doi.org/10.18653/v1/N19-1144} {Predicting the type and
  target of offensive posts in social media}.
\newblock In \emph{Proceedings of the 2019 Conference of the North {A}merican
  Chapter of the Association for Computational Linguistics: Human Language
  Technologies, Volume 1 (Long and Short Papers)}, pages 1415--1420,
  Minneapolis, Minnesota. Association for Computational Linguistics.

\bibitem[{Zeng et~al.(2021)Zeng, Qi, Zhou, Zhang, Hou, Zang, Liu, and
  Sun}]{zeng-2020-openattack}
Guoyang Zeng, Fanchao Qi, Qianrui Zhou, Tingji Zhang, Bairu Hou, Yuan Zang,
  Zhiyuan Liu, and Maosong Sun. 2021.
\newblock \href {https://doi.org/10.18653/v1/2021.acl-demo.43} {Open{A}ttack:
  An open-source textual adversarial attack toolkit}.
\newblock In \emph{Proceedings of the 59th Annual Meeting of the Association
  for Computational Linguistics and the 11th International Joint Conference on
  Natural Language Processing: System Demonstrations}, pages 363--371.

\end{thebibliography}
\bibliographystyle{acl_natbib}


\appendix

\section{Explanatory dialogue systems}
\label{app:systems}

Table~\ref{tab:systems} and Table~\ref{tab:systems_explanans} show the range of existing natural language interfaces and conversational agents for explanations.

\begin{table*}[h!]
    \centering
    
    \resizebox{0.7\textwidth}{!}{%

    \begin{tabular}{|r|cccc|}
    \toprule
    & \multicolumn{3}{c}{\colorbox{champagne}{Task data}}
    & {\colorbox{blond}{Model}}
    \\

    \textbf{Implementations}
    & Num & CV & NLP 
    & 
    \\

    \midrule

    \textsc{dr\_ant} \cite{kuzba-biecek-2020-what-would}
    & \present 
    & 
    & 
    & RF
    \\

    \textsc{ERIC} \cite{werner-2020-eric}
    & \present 
    & 
    & 
    & DT
    \\

    \citet{torri-2021-textual-explanations}
    & \present 
    & 
    & 
    & RF
    \\

    \textsc{TalkToModel} \cite{slack-2022-talktomodel}
    & \present 
    & 
    & 
    & RF
    \\

    \textsc{XAGENT} \cite{nguyen-2022-xagent}
    & \present 
    & \present 
    & 
    & RF, CNN
    \\

    \textsc{ConvXAI} \cite{malandri-2022-convxai}
    & \present 
    &
    & 
    & DT, RF
    \\

    \textsc{ConvXAI} \cite{shen-2023-convxai}
    & 
    & 
    & \data{CODA-19} 
    & Tf 
    \\

    \midrule

    & 
    & 
    & \data{BoolQ}
    & 
    \\

    \textbf{\ilang} (ours)
    & 
    & 
    & \data{DailyDialog}
    & Tf
    \\

    & 
    & 
    & \data{OLID}
    & 
    \\
    
    \bottomrule
    
    \end{tabular} }
    \caption{Explananda (Task and model) comparison of existing implementations of natural language interfaces and conversational agents for XAI.
    We can see that applications to NLP tasks have started to surface only recently. 
    \colorbox{champagne}{Task data} Num = Numeric/Tabular. CV = Computer vision.
    \colorbox{blond}{Explained model} AOG = And-Or graph. \mbox{DT = Decision Tree.} RF = Random Forest. CNN = Convolutional neural network. Tf = Transformer.
    }
    \label{tab:systems}
\end{table*}

\begin{table*}[h!]
    \centering
    \resizebox{\textwidth}{!}{%

    \begin{tabular}{|r|ccccc|cccc|cc|cc|}

    \toprule
    & \multicolumn{5}{c|}{\colorbox{teagreen}{Explanation types}}
    & \multicolumn{4}{c|}{\colorbox{celeste}{Intent recognition / Parsing of user questions}}
    & \multirow{-1}{*}{\tabrotate{\colorbox{babyblue}{Resp}}}
    & \multirow{-1}{*}{\tabrotate{\colorbox{lavenderblue}{DST}}}
    & \multicolumn{2}{c|}{\colorbox{brilliantlavender}{Evaluation}}
    \\

    \textbf{Implementations}
    & FA & CF & Mt & Sim & RG 
    & Comm & Embeds & Fine-Tuned & Few-Shot 
    & & 
    & Auto & Hum \\

    \midrule

    \citet{kuzba-biecek-2020-what-would}
    & \present & \present & & & 
    & {DiF} & & &
    & {DiF}
    & {DiF}
    & 
    & 
    \\

    \citet{werner-2020-eric}
    & \present & \present & & & 
    & & fastText & &
    & Rule
    &
    & 
    & 
    \\

    \citet{torri-2021-textual-explanations}
    & \present & \present & & & 
    & & & \lm{GPT-2} &
    & Rule
    & 
    & 
    & {Like}
    \\

    \citet{slack-2022-talktomodel}
    & \present & \present & \present & & 
    &  & \lm{MPNet} & \lm{T5} & \lm{GPT-Neo/-J}
    & Rule
    & Rule
    & ExM
    & {Like}
    \\

    \citet{nguyen-2022-xagent}
    & \present & \present & \present & & 
    & & \lm{SimCSE} & &
    & Rule
    & 
    & ExM, F1
    & 
    \\

    \citet{malandri-2022-convxai}
    & \present & \present & \present & & 
    & RASA & & &
    & Rule
    & Rule
    & 
    & {Like}
    \\

    \citet{shen-2023-convxai}
    & \present & \present & \present & \present & 
    & & \lm{SciBERT} & &
    & Rule
    & Rule
    &
    & 
    \\

    \midrule

    \multirow{2}{*}{\textbf{\ilang} (ours)}
    & \multirow{2}{*}{\present} & \multirow{2}{*}{\present} & \multirow{2}{*}{\present} & \multirow{2}{*}{\present} & \multirow{2}{*}{\present}
    & & \multirow{2}{*}{\lm{MPNet}} & \lm{BERT+Adap}, & \multirow{2}{*}{\lm{GPT-Neo}}
    & \multirow{2}{*}{Rule}
    & Rule,
    & \multirow{2}{*}{ExM}
    & \multirow{2}{*}{Like}
    \\

    & & & & & & & 
    & \lm{FLAN-T5}
    & 
    & 
    & Adap
    & 
    &
    \\
    
    \bottomrule
    
    \end{tabular} }
    \caption{Explanans (XAI modules) comparison of existing implementations of natural language interfaces and conversational agents for XAI.
    \colorbox{teagreen}{Explanation types} FA~= Feature Attribution. CF~= Counterfactual Generation. Mt~= Meta information about the model. Sim~= Similar examples. RG~= Rationale generation.
    \colorbox{celeste}{Intent recognition} Comm~= Commercial product (RASA~= RASA NLU; DiF~= Google DialogFlow). Embeds~= Nearest neighbor based on sentence embedding.
    \colorbox{babyblue}{Response generation} / \colorbox{lavenderblue}{Dialogue state tracking} Rule = Rule- and template-based response.
    \colorbox{brilliantlavender}{Evaluation}: \textbf{Auto}mated: ExM = Exact match accuracy. \textbf{Human}: Like~= Likert-scale rating.
    }
    \label{tab:systems_explanans}
\end{table*}


\section{\sys{TalkToModel} operations}
\label{app:ttmops}
Most \ttm operations belonging to their ML, Conversation and Description categories can be trivially adapted. Here, we document the changes:

Due to Transformers being explained instead of the much smaller scikit-learn models, we applied small changes such as pre-computing predictions (similar to the tricks we used for attributions and rationales).

\paragraph{Metadata}
For metadata, we provide an operation following the basic idea of model cards \cite{mitchell-2019-model-cards} which supplies information related to model details, intended use of the model, etc., and, analogously, datasheets \cite{gebru-2021-datasheets} for training/test data documentation. User questions can target specific aspects of this structured information and the system replies in natural language and/or tabular formats.

Table~\ref{tab:ttmops} shows the rest of the \ilang operations not depicted by Table~\ref{tab:operations}.

\begin{table}[ht!]
    \centering

    \resizebox{\columnwidth}{!}{%
        \begin{tabular}{|ll|l|}
        \toprule
        \multirow{2}{*}{\centering \rotatebox[origin=c]{90}{\textbf{Filters}}}
        & \texttt{\textcolor{blue}{filter}(id)}
          & Access single instance by its ID \\
        
        & \texttt{\textcolor{blue}{includes}(token)}
          & Filter instances by token occurrence \\

        \midrule 
        \multirow{5}{*}{\centering \rotatebox[origin=c]{90}{\textbf{Prediction}}}
        & \texttt{\textcolor{blue}{predict}(instance)}* 
            & Get the prediction of the given instance \\
        & \texttt{\textcolor{blue}{predict}(dataset)} 
            & Get the prediction distribution across the dataset\\
        & \texttt{\textcolor{blue}{likelihood}(instance)} 
            & Obtain the given instance's probability for each class\\
        & \texttt{\textcolor{blue}{mistakes}(dataset)} 
            & Count number of wrongly predicted instances\\
        & \texttt{\textcolor{blue}{score}(dataset, metric)} 
            & Determine the relation between predictions and labels\\

        \midrule 
        \multirow{3}{*}{\centering \rotatebox[origin=c]{90}{\textbf{Data}}}
        & \texttt{\textcolor{blue}{show}(list)} 
            & Showcase a list of instance\\
        & \texttt{\textcolor{blue}{countdata}(list)} 
            & Count number of instances within the given list\\
        & \texttt{\textcolor{blue}{label}(dataset)} 
            & Describe the label distribution across the dataset\\

        \midrule 
        \multirow{2}{*}{\centering \rotatebox[origin=c]{90}{\textbf{Meta}}}
        & \texttt{\textcolor{blue}{data}(dataset)} 
            & Information related to training/test data\\
        & \texttt{\textcolor{blue}{model}()} 
            & Metadata of the model\\

        \midrule 
        \multirow{2}{*}{\centering \rotatebox[origin=c]{90}{\textbf{About}}}
        & \texttt{\textcolor{blue}{function}()} 
            & Inform the functionality of the system\\
        & \texttt{\textcolor{blue}{self}()} 
            & Self-introduction\\

        \midrule 
        \multirow{2}{*}{\centering \rotatebox[origin=c]{90}{\textbf{Logic}}}
        & \texttt{\textcolor{blue}{and}(op1, op2)} 
            & Concatenation of multiple operations \\ 
        & \texttt{\textcolor{blue}{or}(op1, op2)} 
            & Selection of multiple filters \\
        
        \bottomrule
        \end{tabular}
    }
    \caption{\ttm operations used in \ilang. 
    *Prediction operation provides support for custom input instances received from users.
    }
    \label{tab:ttmops}
\end{table}

\section{Label distributions of NLP use cases}
\begin{figure}
  \centering

  \begin{subfigure}{\columnwidth}
    \centering
    \centering
\includegraphics[width=\linewidth]{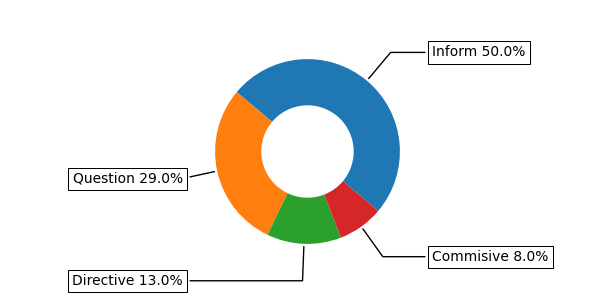}
\caption{Dialogue Act Distribution}
\label{fig:da-distr}
  \end{subfigure}
  \hfill
  \begin{subfigure}{\columnwidth}
\centering
\includegraphics[width=\linewidth]{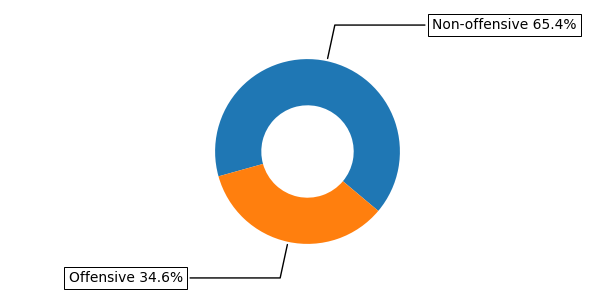}
\caption{OLID Distribution.}
\label{fig:olid-distr}

  \end{subfigure}
  \hfill
  \begin{subfigure}{\columnwidth}
    \centering
    \includegraphics[width=\textwidth]{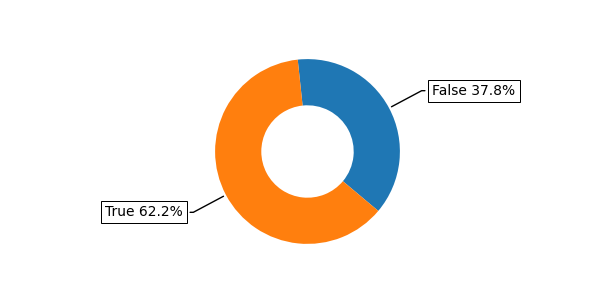}
    \caption{BoolQ Distribution}
    \label{fig:boolq-distr}
  \end{subfigure}

  \caption{Label distribution of all three datasets.}
  \label{fig:subfigures}
\end{figure}

Figure~\ref{fig:subfigures} shows the label distributions of \data{DailyDialog}, \data{OLID} and \data{BoolQ}.

\section{Adapter training details}

Table~\ref{app:adapters-training-params} shows the hyperparameters and training time for the Adapter models for dialogue act classification and slot tagging.

\begin{table*}[h]
    \centering
    \begin{tabular}{lcc}
    \hline
    \textbf{Parameters} & \textbf{Dialogue Act Classification} & \textbf{Slot Tagging}\\
    \hline
    Base Model & \textit{bert-base-uncased} & \textit{bert-base-uncased} \\
    Learning Rate & 1e-4 & 1e-3 \\
    Number of Epochs & 10 & 8 \\
    Batch Size & 32 & 32 \\
    Optimizer & AdamW & AdamW \\
    Number of Labels & 23 & 15 \\
    Avg. Training Time & 53 min & 32 min \\
    Avg. Model Size & 3.6MB & 3.6MB \\
    Training Set & 39,635 & 3,810 \\
    Development Set & 11,010 & 635 \\
    \hline
    \end{tabular}
    \caption{\label{app:adapters-training-params}
    Training parameters for the Adapter-based parsing models. The best performing model was selected based on the loss on the development set. All samples are based on the original prompts automatically augmented through the slot value replacements.}
\end{table*}


\section{Interface}
\label{sec:interface}

We extend the \ttm interface \cite{slack-2022-talktomodel} in the following ways:
\begin{itemize}[noitemsep,topsep=0pt,leftmargin=*]
    \item \textbf{Custom inputs}: Compared to TTM, which only allows user to use instances from three pre-defined datasets, we provide a selection box that allows individual inputs from the user to be considered.
    \item \textbf{Text search}: A search engine that allows the user to filter the dataset according to strings. If a query is present, subsequent operations will consider the subset where this filter is applicable.
    \item \textbf{Dataset viewer}: This shows the first ten instances of the dataset (their IDs and the contents of the text fields) at the start, but in order to make the navigation through the data easier for the user, it will update according to both string filters and operations like label filters.
\end{itemize}

\begin{figure*}[h!]
\centering
\resizebox{0.9\textwidth}{!}{
\begin{minipage}{\textwidth}
\includegraphics[width=\linewidth]{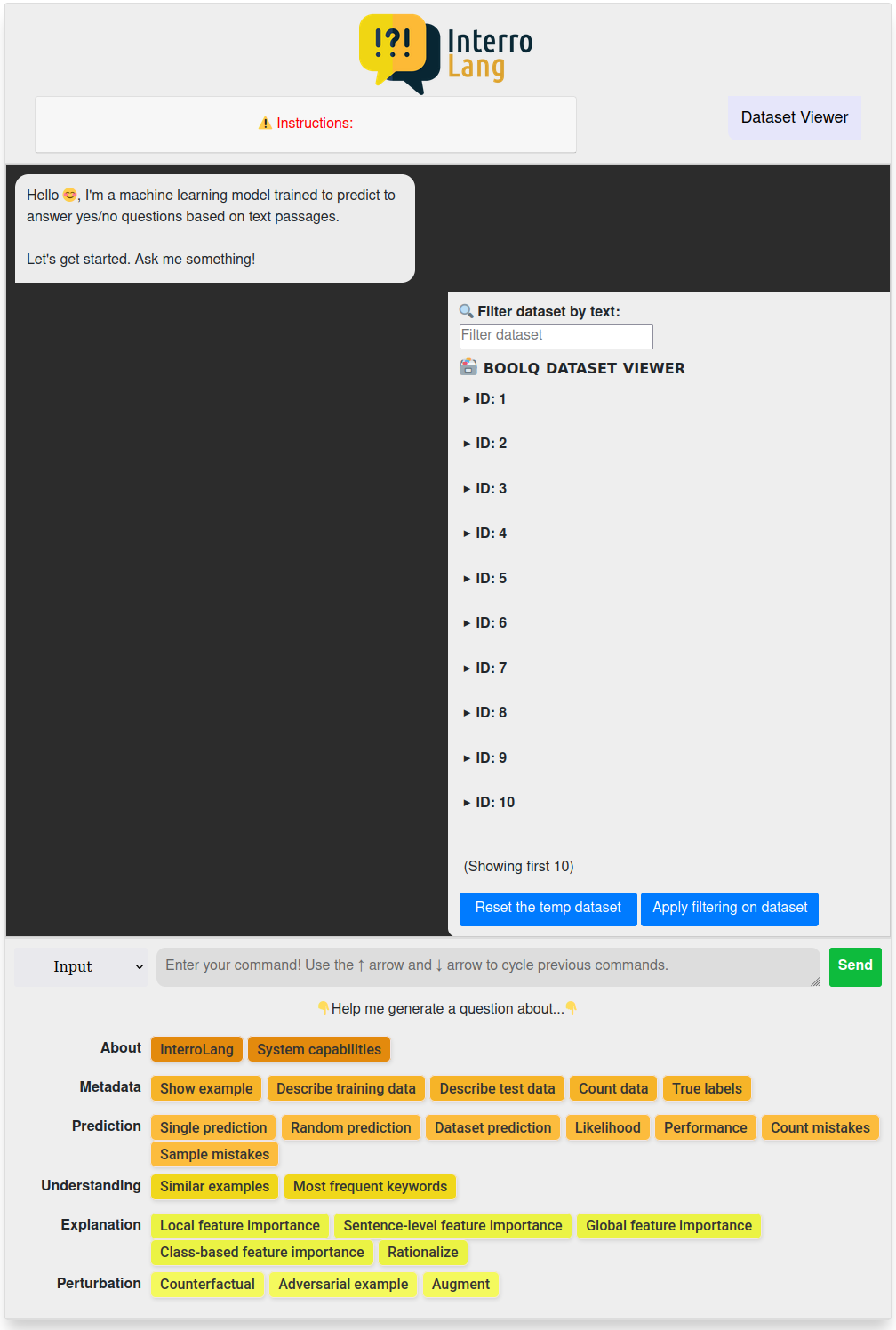}
\end{minipage}
}
\caption{\ilang interface with initial welcome message, opened dataset viewer (\data{BoolQ}) and sample generator buttons.}
\label{fig:interface}
\end{figure*}

\section{Annotation instructions}
\subsection{Task A}
\label{app:instructions-a}
\begin{figure*}[h!]
\centering
\includegraphics[width=\textwidth]{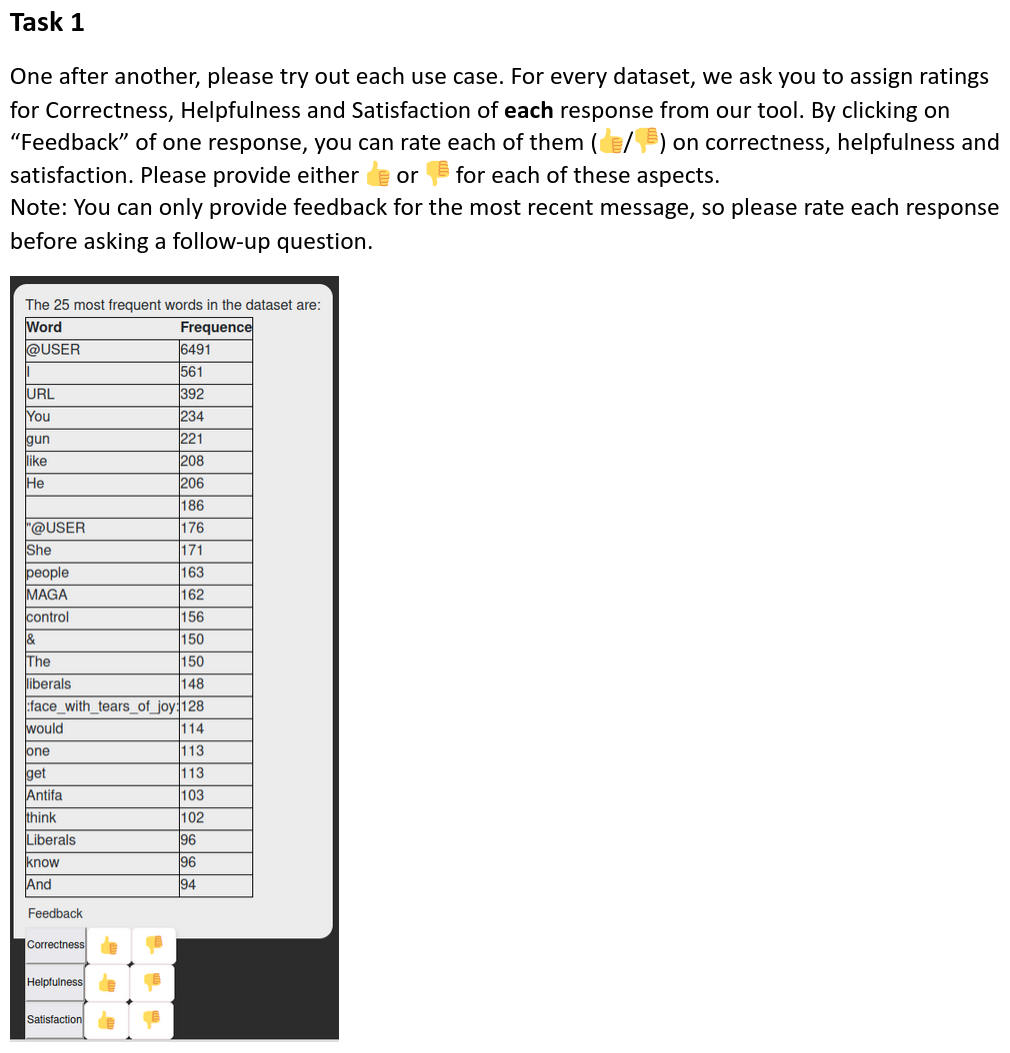}
\caption{User study Task A1: Instruction.}
\label{fig:ilang-study-a1}
\end{figure*}

\begin{figure*}[h!]
\centering
\includegraphics[width=\textwidth]{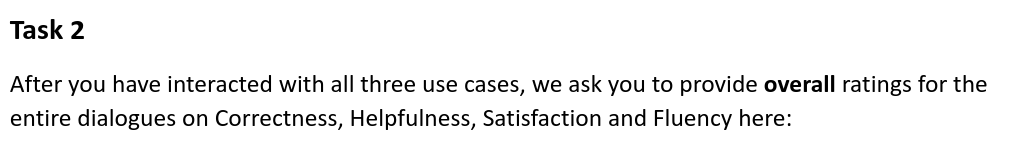}
\caption{User study Task A2: Instruction.}
\label{fig:ilang-study-a2a}
\end{figure*}

\begin{figure*}[h!]
\centering
\includegraphics[width=\textwidth]{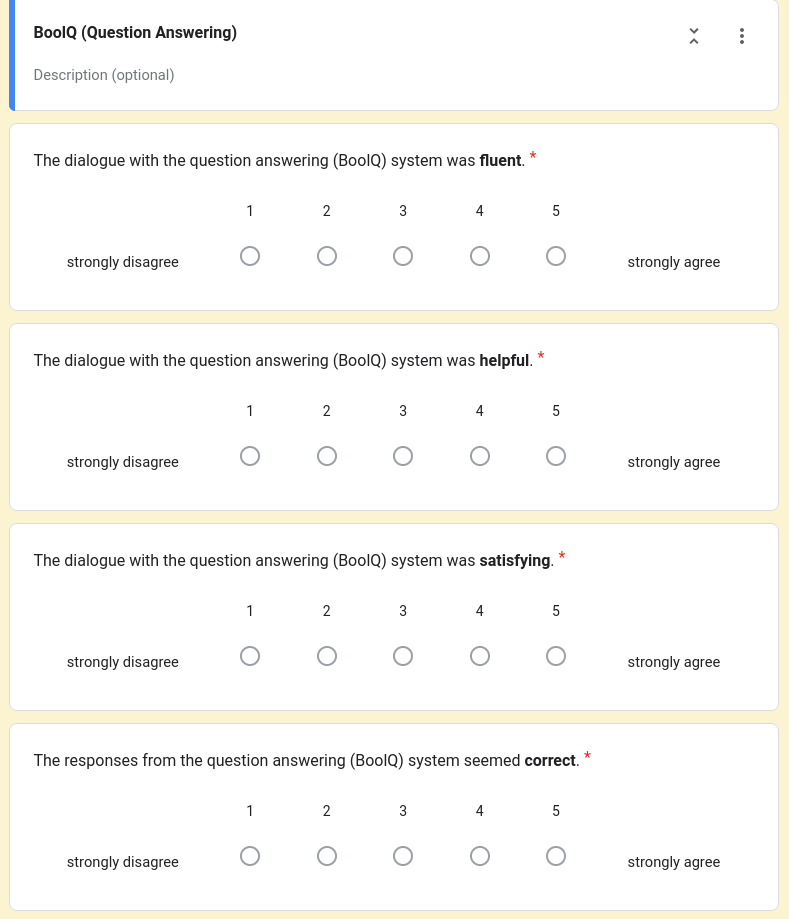}
\caption{User study Task A2: Questionnaire on \data{BoolQ}.}
\label{fig:ilang-study-a2b}
\end{figure*}

Figure~\ref{fig:ilang-study-a1} and Figure~\ref{fig:ilang-study-a2a} show the instructions of the user study on subjective ratings (Task A) as described in \S \ref{sec:evaluation-task-a}. Figure~\ref{fig:ilang-study-a2b} shows a screenshot of the Google Forms in Task A2.

\subsection{Task B}
\label{app:instructions-b}
\begin{figure*}[h!]
\centering
\includegraphics[width=\textwidth]{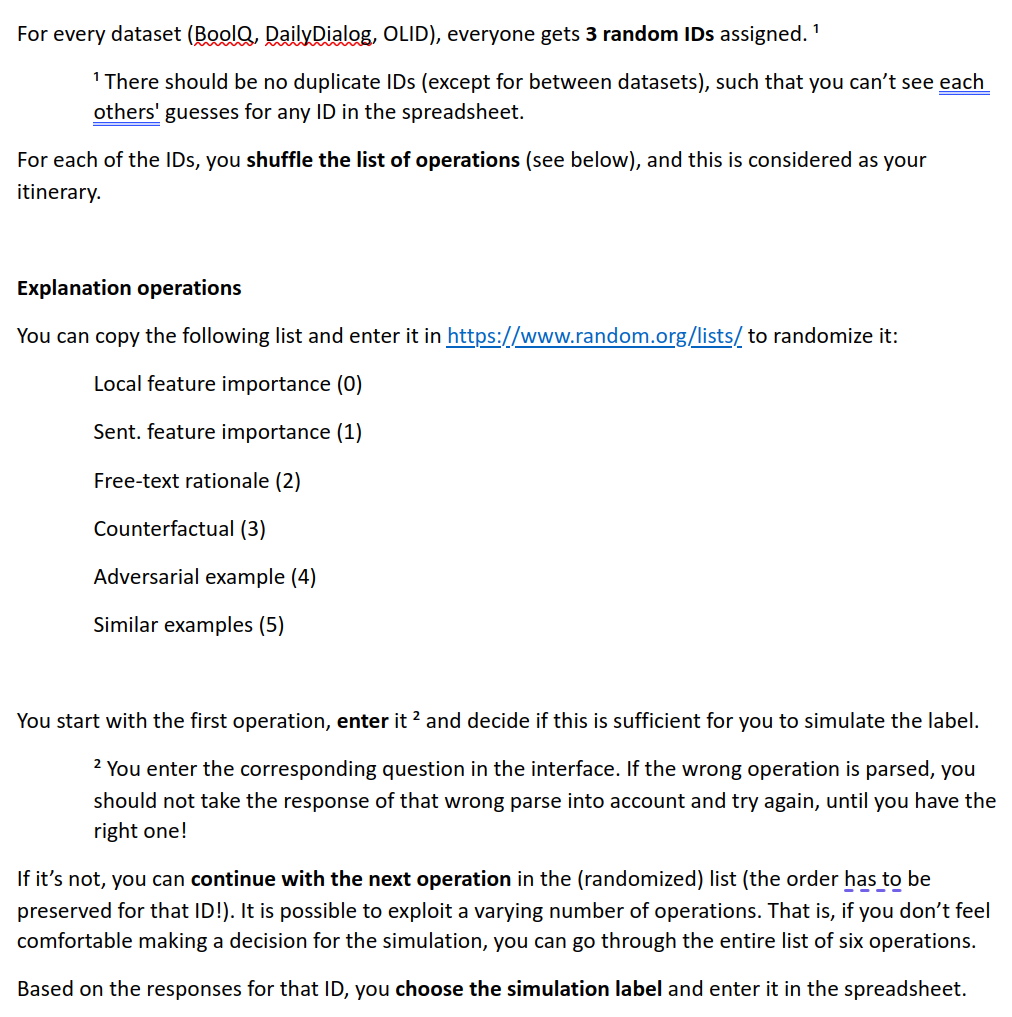}
\caption{User study Task B.}
\label{fig:ilang-study-b}
\end{figure*}

Figure~\ref{fig:ilang-study-b} shows the instructions of the user study on simulatability described in \S \ref{sec:evaluation-task-b}.

\section{\ilang Dataset statistics}
\label{app:dataset}

Across all three datasets we have 659 unique user questions that don't overlap with the \ilang sample prompts (81.16\%) and 153 questions that do overlap. The high number indicates that our prompts approximate the actual user questions rather well. On the other hand, some of the user questions were taken directly from the prompt examples.

In particular, \data{OLID} has 180 (61.2\%) unique user questions with 114 overlaps; \data{DailyDialog} has 208 (69.3\%) unique user questions and 92 overlaps; \data{BoolQ} has 192 (88.1\%) unique user questions, 26 overlaps.
Across all three datasets this results in 478 unique questions (58.9\%) and 334 overlapping ones.

\clearpage

\begin{table*}[hbt!]
    \centering
    \section{Sample Dialogue (\data{BoolQ} with Adapter)}\label{app:dialogue-example}
    \begin{tabular}[t]{p{6.5cm}p{8cm}}
        \includegraphics[width=1\linewidth]{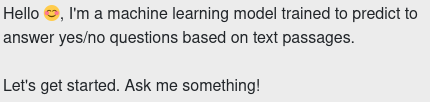} &\\
        User: Hi! Which kind of a model do you use? & \includegraphics[valign=t,width=1\linewidth]{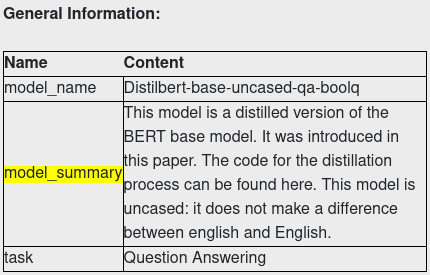} \\
        User: And what is the dataset?
         & \includegraphics[valign=t,width=1\linewidth]{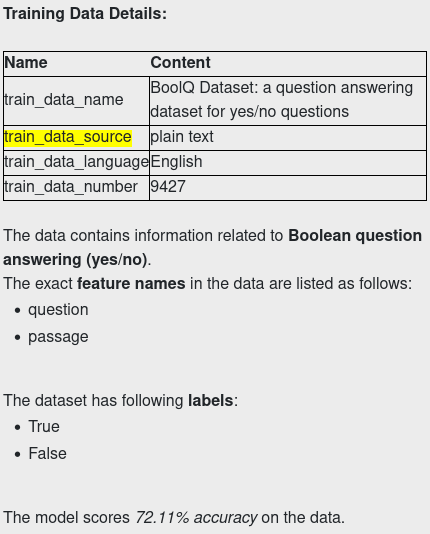} \\
        User: Can you show me how often the model makes incorrect predictions? & \includegraphics[valign=t,width=1\linewidth]{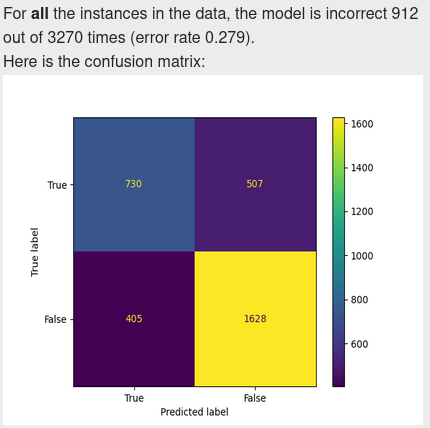}\\
    \end{tabular}
\end{table*}
\begin{table*}[hbt!]
    \centering
    \begin{tabular}[t]{p{6.5cm}p{8cm}}
        User: Can you show me some examples of the mistakes? & \includegraphics[valign=t,width=1\linewidth]{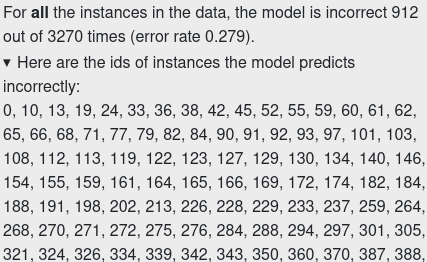} \\
        User: Ok, great! What about the id 42? Can you show me this sample? & \includegraphics[valign=t,width=1\linewidth]{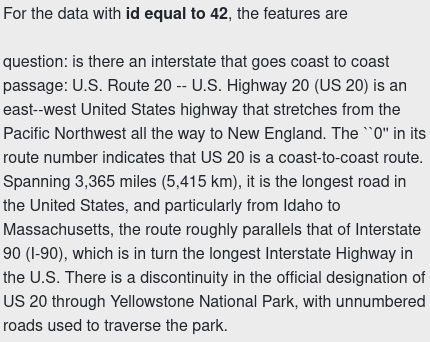}\\
        User: Please show me the token attributions for this id. & \includegraphics[valign=t,width=1\linewidth]{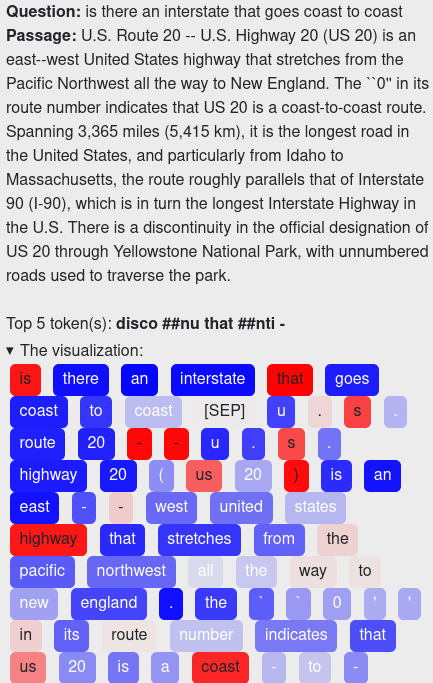}\\        
    \end{tabular}
    \label{tab:my_label}
\end{table*}

\clearpage

\begin{table*}[]
    \centering
    \begin{tabular}{p{6.5cm}p{8cm}}
        User: Can you do an adversarial attack?  &
        \includegraphics[valign=t,width=1\linewidth]{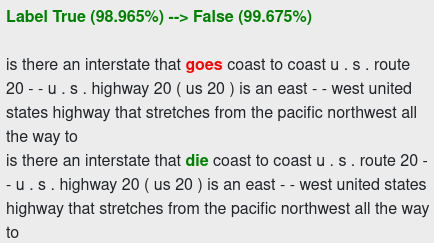} \\
        User: Ok, thanks! Looks good :)  & \includegraphics[valign=t,width=0.35\linewidth]{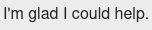} \\
        User: What would be the counterfactual for this instance?  & \includegraphics[valign=t,width=1\linewidth]{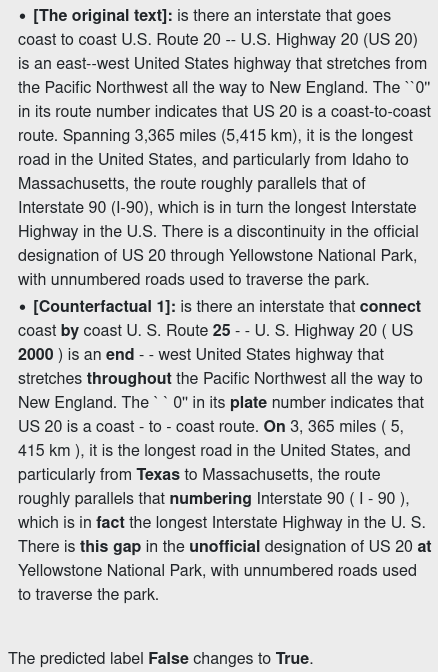} \\
        User: Can you show me the most important features overall (across all data )?  & \includegraphics[valign=t,width=1\linewidth]{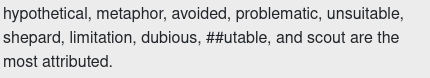}   \\
        User: Ok, I think that's it for today. Bye!  & \includegraphics[valign=t,width=0.35\linewidth]{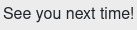}   \\
    \end{tabular}
    \label{tab:dialog-sample}
\end{table*}

\end{document}